\pdfoutput=1

\documentclass[11pt]{article}

\usepackage[final]{acl}

\usepackage{times}
\usepackage{latexsym}

\usepackage[T1]{fontenc}

\usepackage[utf8]{inputenc}

\usepackage{microtype}

\usepackage{xcolor}

\usepackage{soul}
\usepackage{url}
\usepackage[hidelinks]{}
\usepackage[utf8]{inputenc}
\usepackage{caption}
\usepackage{graphicx}
\usepackage{amsmath}
\usepackage{booktabs}
\usepackage{tikz}
\tikzset{every picture/.style={remember picture}}
\usetikzlibrary{tikzmark}
\usepackage{multirow}
\usepackage{makecell}
\usepackage{hhline}
\usepackage[normalem]{ulem}
\usepackage{paralist}
\usepackage{hyperref}
\usepackage{float}
\usepackage[normalem]{ulem}

\usepackage{soul}
\usepackage{url}
\usepackage[hidelinks]{}
\usepackage[utf8]{inputenc}
\usepackage{caption}
\usepackage{graphicx}
\usepackage{amsmath}
\usepackage{booktabs}
\usepackage{tikz}
\tikzset{every picture/.style={remember picture}}
\usetikzlibrary{tikzmark}
\usepackage{multirow}
\usepackage{makecell}
\usepackage{hhline}
\usepackage{paralist}
\usepackage{amssymb}
\usepackage{subcaption}
\usepackage[ruled,vlined]{algorithm2e}

\usepackage{graphicx,calc}
\newlength\myheight
\newlength\mydepth
\settototalheight\myheight{Xygp}
\newcommand*\inlinegraphics[1]{%
  \settototalheight\myheight{Xygp}%
  \settodepth\mydepth{Xygp}%
  \raisebox{-\mydepth}{\includegraphics[height=\myheight]{#1}}%
}

\usepackage{tikz}
\usepackage{enumitem}
\definecolor{brightcerulean}{rgb}{0.11, 0.67, 0.84}

\newcommand{\sysname}{StoRM}

\usepackage{amssymb,amsmath,amsthm,enumitem}

\usepackage{subcaption}
\DeclareCaptionSubType*[Alph]{table}
\DeclareCaptionLabelFormat{mystyle}{Table~\bothIfFirst{#1}{ ̃}#2}
\title{Guiding Neural Story Generation with Reader Models}

\author{Xiangyu Peng, Kaige Xie, Amal Alabdulkarim, Harshith Kayam, Samihan Dani, Mark O. Riedl \\
  Georgia Institute of Technology \\
  \texttt{\{xpeng62,kaigexie,amal,hkayam3,sdani30,riedl\}@gatech.edu} \\}

\begin{document}
\maketitle

\begin{abstract}
Automated storytelling has long captured the attention of researchers for the ubiquity of narratives in everyday life. 
However, it is challenging to maintain coherence and stay on-topic toward a specific ending when generating narratives with neural language models. 
In this paper, we introduce Story generation with Reader Models (\sysname), a framework in which a \textit{reader model}
is used to reason about the story should progress.
A reader model infers what a human reader believes about the concepts, entities, and relations about the fictional story world.
We show how an explicit reader model represented as a knowledge graph
affords story coherence and provides controllability in the form of achieving a given story world state goal.
Experiments show that our model produces significantly more coherent and on-topic stories, outperforming baselines in dimensions including plot plausibility and staying on topic. 
\end{abstract}

\section{Introduction}
\begin{figure}[t!]
    \centering
    \includegraphics[width=\linewidth]{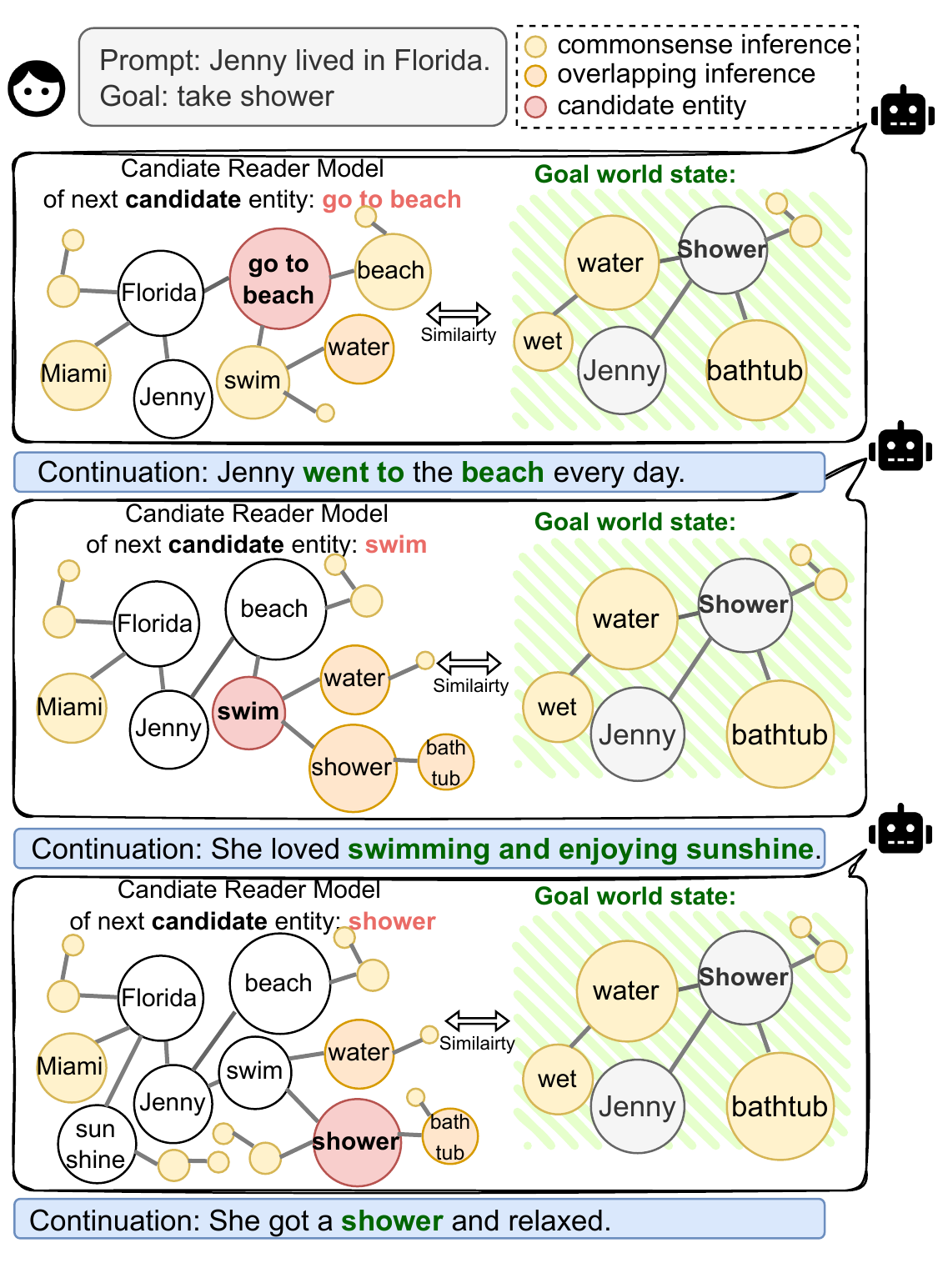}
    \caption{The overview of StoRM system. 
    Our goal is to generate a story on a prompt for reaching the given goal.
    1. The system builds a goal world state \inlinegraphics{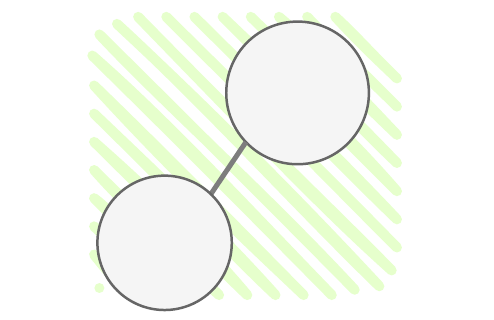} by converting natural language text into knowledge graph and then expands it with commonsense inference \inlinegraphics{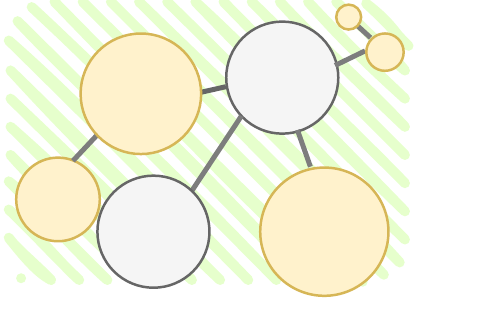}. 
    2. \sysname{} builds a prompt story world \inlinegraphics{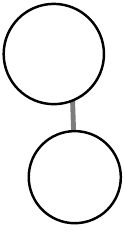} and then infers a set of concepts \inlinegraphics{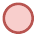} on each entity in prompt story world. 
    3. For every concept \inlinegraphics{figures/inference_node.pdf}, \sysname{} obtains a candidate reader model \inlinegraphics{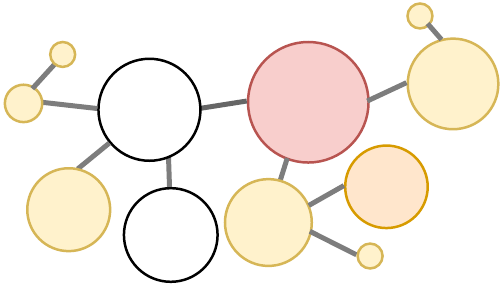} by inferring commonsense relations on it. 
    4. \sysname{} chooses the \textit{topk} concepts which maximize its similarity with goal world state.
    5. A language model generates continuation options on the \textit{topk} concepts, which are added to the story. }
    \label{fig:figure1}
\end{figure}

Automated Story Generation is the challenge of designing an artificial intelligence system that can generate a natural language text that is perceived by readers as a story.
Early work on story generation used symbolic planning~\cite{meehan1976metanovel,lebowitz1987planning,cavazza2003interacting,porteous2009controlling,riedl2010narrative,ware2010modeling,ware2021}.
These systems would be provided with a description of the initial world state---usually a list of predicates---and a goal---a description of what predicates should be true to be successful.
These approaches had two benefits.
First, the plots tended to be coherent because of logical constraints on the actions. %
Second, the plots were guaranteed to end in a state in which the goal held.
However, these systems require substantial knowledge engineering of logical constraints, limiting their generality, and don't always generate plot or stories in natural language.
\begin{table*}[!tbh]
\footnotesize
\centering
\setlength\tabcolsep{2.5pt} %
\begin{tabular}{p{0.01\linewidth}|p{0.16\linewidth}|p{0.36\linewidth}|p{0.43\linewidth}}
    \toprule
    & ROCStories & Writing Prompts & Fairytale\\
    \midrule
    \parbox[t]{2mm}{\multirow{3}{*}{\rotatebox[origin=c]{90}{SToRM}}}
    &\textbf{Charles had always wanted to have a college degree.} 
    
    He was admitted to a college. 
    
    Finally, he graduated from college. 
    
    He was glad he finally \textbf{\uwave{got his diploma}}.
    &
     \textbf{Jennifer wander in hallway.}
    She thought she had found something on the ground. When she looked at the ground, there was nothing. 
 She then kept walking in the dark.
 She cannot see any light.
    
    Then, \textbf{she fell on ground}.
    She was in a lot of pain.
    When she got back up, she fell again. 
    This time, she got back up and just walked away. 
    ``Hello?'' she asked.
    
    \textbf{She walked toward kitchen.}
    &\textbf{Hero consumed them.} Then, he got very sick and died. He buried himself. Then, he reborned.
    \textbf{\uwave{He showed up.}} He made a speech to the people. He received applause from the audience. After, he thanked the audience for their support. When he finally came back, everyone cheered. 
    \textbf{Hero invoked command to his magic sword.} Immediately, he became more powerful than ever. Now, he was the boss of the world. He began to be in charge of the world. He would be a king. \textbf{He had again assumed his place as King of Golden Mountain.}\\
    \midrule
    \parbox[t]{2mm}{\multirow{3}{*}{\rotatebox[origin=c]{90}{Baselines (C2PO or CP)}}} &
    \textbf{{\color{red}{[C2PO]}} Charles had always wanted to have a college degree.} 
    Charles begins get a job. 
    Charles tries to work hard. 
    Charles wants to get a degree.
    \textbf{Charles was proud that he had a college degree.}
    &
    \textbf{{\color{red}{[CP]}} Jennifer wander in hallway.}``Hey,'' said the woman in front of her. ``Is this the elevator? I think it is.'' ``I don't think of that. It looks like its the elevator.'' ``Jennifer asked you 're the only thing keeping the girl alive'', The girl said.
    \textbf{Jennifer fell on ground.} She was on her way home from work when she saw a strange figure in the kitchen. She turned. ``Hey.'' Jennifer said to an odd looking man. All he saw was a girl that looked like it that looked back at him. \textbf{Jennifer walked toward kitchen.}
    &\textbf{{\color{red}{[C2PO]}} Hero consume them.} Hero starts to eat something. Hero wants drink water. Hero begins to talk to someone. Hero begins to have a reason.
    
    \textbf{Hero revealed himself.} Hero starts to tell everyone about it. Hero wants to practice. Hero tries to know magic.
    
    \textbf{Hero invoked command to his magic sword.} Hero wants to prove his skills. Hero wants to show off. Hero wants to be in charge. Hero wants to be king. \textbf{He had again assumed his place as King of Golden Mountain.}\\

    \bottomrule
\end{tabular}
\caption{Story examples generated by \sysname{}, C2PO, and \citet{goldfarb2020content} (CP). Stories generated on ROCStories and Fairytale stories are by C2PO and the story on Writing Prompts is generated by \citet{goldfarb2020content}. 
\textit{Bolded} text are prompts and goal texts.  \uwave{Underlined} text indicates early stopping when the generated story hits the goal.
More details about generation can be found in Appendix~\ref{app:exp_coherence}.}
\label{tab:examples}
\end{table*}

Recently, 
neural language modeling approaches \cite{roemmele2016writing,khalifa2017deeptingle,martin2018event,clark2018,yao2019plan, rashkin2020plotmachines, fan2019strategies, ammanabrolu2020automated} have been applied to story generation because they circumvent the need for manual knowledge engineering and tend to produce relatively fluent, varied, and naturalistic language.
Language models are, however, not goal-directed. 
That is, one cannot natively provide both a context prompt and a goal to be achieved after an arbitrary number of continuations. 
Further, language models struggle with maintaining story coherence---the logical progression of events---and may also become repetitive.
Large, pre-trained language models improve fluency and generalization but do not provide goal-directedness and stories generated can still be perceived as lacking in coherence in the sense that they meander without direction.

We consider the challenge of coherent and controllable text generation for neural language model based story generation.
We hypothesize that neural language models, while powerful text-completion systems, are not natively well-suited for coherent story generation because a neural network trained with a cross-entropy loss function is unlikely to model the unfolding context of a story the same way as a human reader.
Studies of human reader comprehension~\cite{zwaan1998situation} 
show that readers comprehend stories by tracking the relations between entities and events in ways that can be expressed as a graph.
The perceived coherence of a story is a function of the connectedness of this graph~\cite{graesser1994constructing}. 
Ensuring the causality between sentences can improve the coherence of stories \cite{peng2021inferring}.

Inspired by cognitive science, we aim to augment neural language models with a {\em\bf reader model} in which a story generation system infers a graph of concepts, entities, and relations that a reader is likely to believe about the story world as they read an incrementally generated story.
The reader model enables the story generation algorithm to explicitly reason about the entities and relations and generate story continuations that use those entities to move the story forward; a reader can track how entities and relations change over time and thus perceive stories as more coherent.
We use large language models to produce the continuation text of the story generation.
However instead of only providing the previous story as context, our algorithm also selects one or more entities from the world model and uses template filling to generate candidate continuations.

The reader model also provides a means for directing the generation process.
In addition to a starting context prompt, we require a goal to be given.
The goal is natural language text which is then converted into 
a knowledge graph 
(See Fig.~\ref{fig:figure1}).
The goal knowledge graph provides a rough outline of the entities and relations that need to be present in the story but without providing particulars about everything that must be in the story or the ordering in which they must occur.

Our contributions are as twofold: (1) we propose an automated story generation model with Reader Models (\textbf{\sysname}) which maintain coherence and controllability of generated stories at the same time; and (2) we conduct a thorough experimental study against strong baselines which shows that
\sysname{} produces significantly more coherent and goal-directed story.

\section{Related Work and Background}

We situate our paper in the literature of neural networks---recurrent and transformer-based---to produce stories~\cite{roemmele2016writing,khalifa2017deeptingle,martin2018event,clark2018, fan2018hierarchical}.
There are a few works that are highly related to our proposed framework, in terms of the following two dimensions: the generation controllability and the usage of commonsense knowledge.
The controllability in story generation focuses on how to enable the generation process to adhere to the user's inputs.
\citeauthor{goldfarb2020content}~\shortcite{goldfarb2020content} conducts generation in two steps: planning a story plots based on a prompt, then generating a story by filling the mask tokens in plots with BART\cite{lewis2020bart}.
Plot Machines~\cite{rashkin2020plotmachines} accepts as an input an un-ordered outline of concepts and conditions a language model.

Commonsense knowledge plays an important role in story generation.
The most popular way of utilizing it is to train neural language models (e.g. GPT-2 \cite{radford2019language}) on commonsense knowledge bases such as ConceptNet~\cite{speer2013conceptnet} and ATOMIC~\cite{sap2019atomic,Hwang2021COMETATOMIC2O} which contains detailed information regarding well-known facts or causal relationships.
Thus the resulting language model, named COMET~\cite{bosselut2019comet, Hwang2021COMETATOMIC2O}, becomes capable of inferring new commonsense knowledge on novel phrases.
Given a sentence, COMET infers commonsense attributes about the characters in three categories:     
(1) social interactions (i.e. \texttt{xNeed} indicates what the character of this sentence needed),
(2) physical entities, and
(3) effect of events inferred from the sentence.
~\citet{ammanabrolu2021automated} proposes Causal, Commonsense Plot Ordering (C2PO) framework which takes advantage of COMET to infer predecessor and successor events and then bi-directionally search from pre-specified start event to end event, however, C2PO generates plots made up of highly constrained, templated text; ~\citet{peng2021inferring} leverages COMET to infer the character intentions and effects of actions so as to guide the generation process, but they did not consider controllability.
There are also other approaches that directly incorporate commonsense knowledge graphs into the encoding process \cite{mihaylov2018knowledgeable, guan2019story}.
Compared with them, the novelty of our paper is to improve coherence and controllability at the same time with the help of external commonsense knowledge graph.

\section{Story Generation with Reader Models}
\label{sec:rm}

\begin{figure*}[!t]
    \centering
    \includegraphics[width=\textwidth]{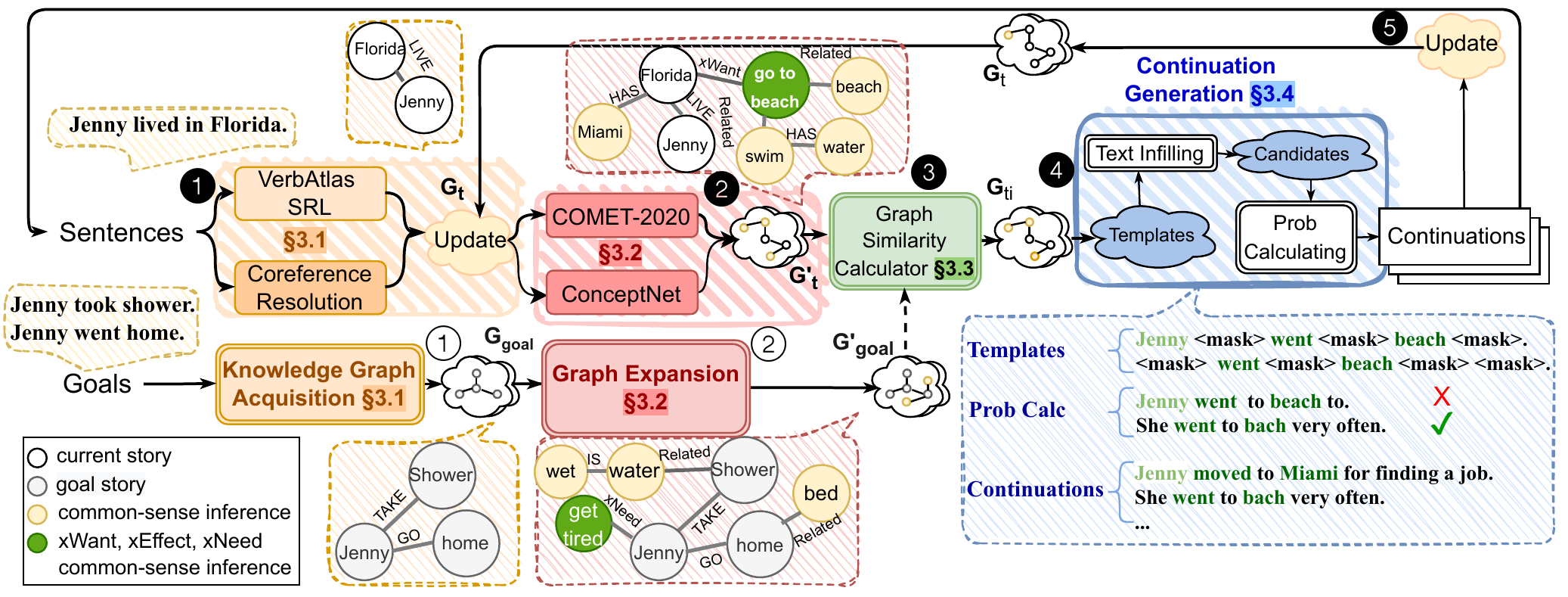}
    \caption{The overall procedure of \sysname.
    \inlinegraphics{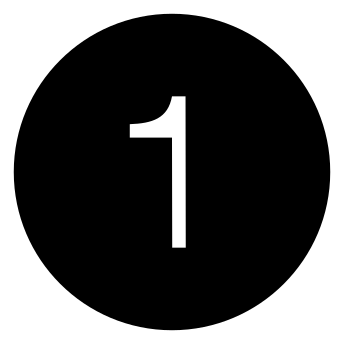} \inlinegraphics{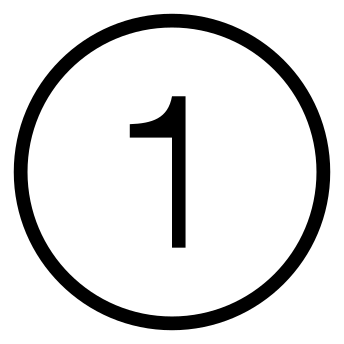} 
    Prompts and goals are transformed into knowledge graph (\S\ref{sec:kg}).
    \inlinegraphics{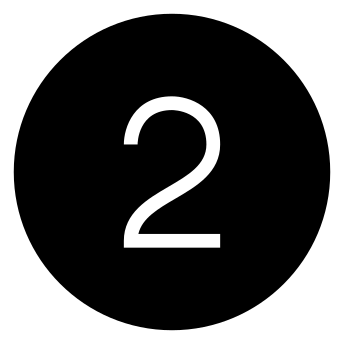}
    \inlinegraphics{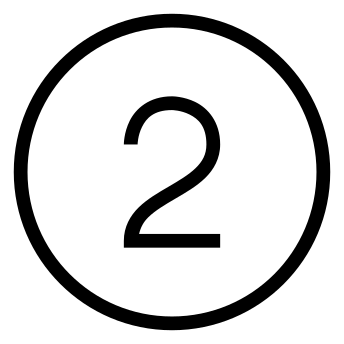} They are expanded with inferences (\S\ref{sec:infer}) and obtain a set of candidate knowledge graphs $\mathbf{G}'_t$.
    \inlinegraphics{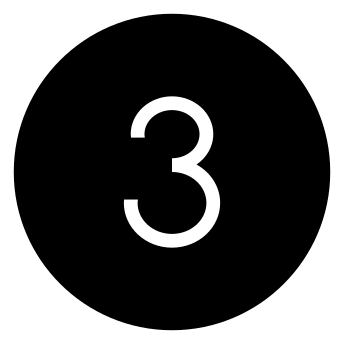}
    $\mathbf{G}'_t$ are compared to 
    and \textit{topk} candidate knowledge graphs are kept (\S\ref{sec:graph}).
    \inlinegraphics{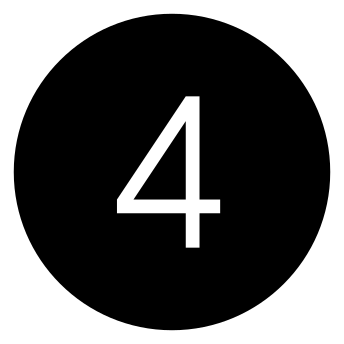}
    Generate story continuations on \textit{topk} candidate knowledge graphs (\S\ref{sec:gen}).
    \inlinegraphics{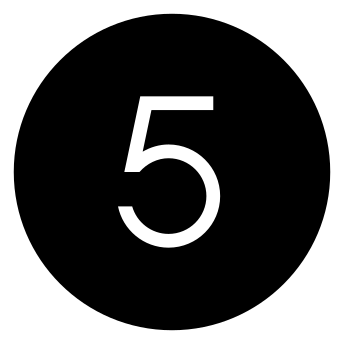}
    Story continuations are appended to story history and also update candidate knowledge graphs with details in continuations.
    }
    \label{fig:pipeline}
\end{figure*}

In this section, we introduce a framework---{\em Story generation with Reader Models} (\sysname)---for generating stories with 
models of what the reader will believe about the fictional story world.
We hypothesize that the incorporation of a {\em reader model} into the story generation process will increase story coherence. 
We define {\em story coherence} as the extent to which readers can identify connections between different events and entities in a story. 
In this work, the reader model is represented as a {\em knowledge graph}, a set of triples of the form $\langle subject, relation, object\rangle$.
By making the beliefs about what the reader likely knows explicit, we provide mechanisms for selecting which entities to include in the continuation of the story.
Because the \sysname{} framework maintains a knowledge graph that approximates the reader's beliefs about the story world, we are able to compare the reader model to a desired goal world state, also described as a knowledge graph.
The \sysname{} framework is thus {\em controllable}.

Our framework starts with a prompt and a set of goals
(See Fig.~\ref{fig:pipeline}), which are natural language texts.
The prompt and goals are transformed into knowledge graphs by extracting entities (\S\ref{sec:kg}), referred as reader model and goal world state (see yellow bubbles in Fig.~\ref{fig:pipeline}).
These two knowledge graphs are expanded using commonsense techniques (\S\ref{sec:infer}; See red bubbles in Fig.~\ref{fig:pipeline}). 
For every inferred entity in the reader model, we further expand the reader model with its commonsense inference to obtain a candidate knowledge graph, which is scored based on how similar it is relative to the goal world state (\S\ref{sec:graph}).
Entities of which the candidate KG shares the \textit{topk} similarity with the goal world state are selected and 
the generation technique uses templates to generate possible story continuations (\S\ref{sec:gen}; See blue bubbles in Fig.~\ref{fig:pipeline}).
By targeting different entities and using template infilling, we reduce neural network hallucination of new entities and create a diverse set of story continuations.
The selected continuation starts the next iteration of the generation process.
Story examples can be found in Table~\ref{tab:examples}.

\subsection{Knowledge Graph Acquisition}
\label{sec:kg}
With the automatic generation of the story, some important information could be forgotten. 
The knowledge graph is an explicit and {\em persistent} memory of entities mentioned or inferred from the story text generated so far.
Knowledge Graphs represent information in the form of triples, consisting of
a subject entity, relation and object entity. 
For example, ``\textit{Jenny lived in Florida}'' is represented as
$\langle jenny, live, florida\rangle$.
The entities represent the nodes of the graph and their relations act as edges. 

To acquire the knowledge graph,
we firstly trained a Semantic Role Labeling (SRL) model \cite{gildea2002automatic} on VerbAtlas \cite{di2019verbatlas}---a hand-crafted
lexical-semantic resource whose goal is to bring together all verbal synsets from WordNet \cite{fellbaum1998wordnet} into semantically-coherent frames.
This SRL model provides the automatic identification and labeling of argument structures of stories.
Further detail of training can be found in Appendix~\ref{sec:verbatlas-srl}.

\sysname{} then converts the output of VerbAtlas SRL model into knowledge graph triples.
Entities represent the theme and attribute and VerbAtlas frames act as edges. 
An example is shown in yellow bubbles of Fig.~\ref{fig:pipeline}.
Multiple character names, object names and pronouns make the knowledge graph representation hard to interpret. 
Hence, we adopt a end-to-end Coreference Resolution model \cite{lee2017end} to find all expressions that refer to the same entity in a story to minimize the entities.

\sysname{} starts with two knowledge graphs.
The first, $\mathbf{G}_1$, is the converted 
prompt (first sentence). 
The second $\mathbf{G}_{\rm goal}$ is the converted goal.
With the generation of the continuation candidates, we will update the knowledge graph $\mathbf{G}_t$ with new continuations to get new knowledge graph $\mathbf{G}_{t+1}$, where $t$ is the index of the sentence in the story.
Because we obtain the \textit{topk} continuation knowledge graphs, hence $\mathbf{G}_t = \{\mathbf{G}_{t1}, ..., \mathbf{G}_{tK}\}$.

\subsection{Graph Expansion}
\label{sec:infer}
Human readers use commonsense knowledge to infer the presence of entities and concepts not explicitly mentioned in story text. 
For example, ``Florida'' has ``beaches'' and ``eating dinner'' implies ``dishes''.
In accordance, we use common-sense inferences of entities to expand the knowledge graph to provide entities for characters to interact with and thus drive the story forward.
Because the presence of entities and concepts are inferred from prior events, the reader should be able to track the connections between entities and events, thus supporting perceived story coherence.

We consider two types of nodes in knowledge graph---physical entities  (i.e. ``beach'') and social events (i.e. ``go to beach'').
We use ConceptNet5~\cite{speer2013conceptnet}---a multilingual knowledge base, representing words and phrases that people use and the common-sense relations between them---to
infer each physical entity.
We use $\text{COMET}^{20}_{20}$~\cite{Hwang2021COMETATOMIC2O}---a transformer-based generative model trained on the $\text{ATOMIC}^{20}_{20}$ commonsense dataset~\cite{Hwang2021COMETATOMIC2O}---to infer relations on events  about commonsense relations of social interaction.
More details about inference types can be found in Appendix~\ref{app:comet_inference}.

Firstly, we expand the physical entities in current knowledge graph $\mathbf{G}_{ti}$ to obtain a \textit{physical entity candidate set} $E_{ti,{\rm entity}}$ with ConceptNet (i.e. expand ``Florida'' with ``Miami''; yellow nodes in Fig.\ref{fig:pipeline}).
We then expand each entity $e^{k_1}_{ti,{\rm entity}} \in E_{ti,{\rm entity}}$ with ConceptNet until we reach a depth of $m_1$ to obtain $\widetilde{E^{k_1}_{ti,{\rm entity}}}$, where $k_1 \in [1,n_1]$ and $n_1$ is the size of $E_{ti,{\rm entity}}$.
Secondly, we generate social interactions of current story history by $\text{COMET}^{20}_{20}$ to obtain an \textit{social event candidate} set $E_{ti,event}$ (i.e. we infer ``Jenny lived in Florida'' to ``go to beach''; green nodes in Fig.~\ref{fig:pipeline}) . 
We expand each event $e^{k_2}_{ti,event} \in E_{ti,event}$ 
with $\text{COMET}^{20}_{20}$ and ConceptNet until we reach a depth of $m_1$ to obtain $\widetilde{E^{k_2}_{ti,event}}$, where $k_2 \in [1,n_2]$ and $n_2$ is the beam size of the output of $\text{COMET}^{20}_{20}$.
Hence, for $\mathbf{G}_{ti}$, we totally obtain $n$ entity candidates $e_{ti}^k$ $\in$ $\{E_{ti,{\rm entity}}$ $\cup$ $E_{ti,event}\}$, where $k \in [1,n]$ and $n = n_1 + n_2$ and its corresponding
knowledge graph candidates $\mathbf{G}'_{ti} = \{\mathbf{G}^{'1}_{ti}, ..., \mathbf{G}^{'n}_{ti}\}$, where $\mathbf{G}^{'k}_{ti} = \mathbf{G}_{ti} \cup \widetilde{E^{k}_{ti}}$ 
($\widetilde{E^{k}_{ti}} = \widetilde{E^{k}_{ti,{\rm entity}}}$ for $e^k_{ti} \in E_{ti,{\rm entity}}$; $\widetilde{E^{k}_{ti}} = \widetilde{E^{k}_{ti,{\rm event}}}$ for $e^k_{ti} \in E_{ti,{\rm event}}$).
Totally, we obtain $\mathbf{G}'_t$ $=$ $\{\mathbf{G}'_{t1},$ $...,$ $\mathbf{G}'_{tK}\}$.

We also expand the depth of the graph for goals by $m_2$ via similar means: we firstly expand all the physical entities in $\mathbf{G}_{\rm goal}$ by $m_2$ depth with ConceptNet to obtain inference set $\widetilde{E_{\rm goal, enitity}}$.
Then we expand social interactions of the goal text with $\text{COMET}^{20}_{20}$ and ConceptNet by $m_2$ depth to obtain event candidates $\widetilde{E_{\rm goal,event}}$.
Finally, we obtain the updated knowledge graph $\mathbf{G}'_{goal} = \mathbf{G}_{\rm goal} \cup \widetilde{E_{\rm goal, enitity}} \cup$ $\widetilde{E_{\rm goal,event}}$.
Further details can be found in Appendix~\ref{app:expand}.

\subsection{Graph Similarity}
\label{sec:graph}

We achieve controllability of generated continuations by calculating the \textit{graph similarity} between the candidate knowledge graph $\mathbf{G}^{'k}_{ti}$ and the goal knowledge graph $\mathbf{G}'_{\rm goal}$. The candidate knowledge graph $\mathbf{G}^{k}_{ti}$ is updating $\mathbf{G}_{ti}$ with candidate entity $e^k_{ti}$.

We calculate the knowledge graph similarity score:
\begin{align}
    R(\mathbf{G}^k_{ti}) &=
    \alpha \times r_1(\mathbf{G}^k_{ti}, \mathbf{G}_{\rm goal})  \nonumber\\
          & + (1 - \alpha) \times r_2(\widetilde{E^{k}_{ti}}, \mathbf{G}'_{\rm goal}))
\end{align}
where $r_1$ is story entity overlapping score and $r_2$ is inference overlapping score. 
$\alpha$ is a hyper-parameter to control the inference's contribution on calculating overlapping rate.
$\widetilde{E^{k}_{ti}}$ is inference set of candiate entity $e^k_{ti}$.   

\textit{Story entity overlapping score} ($r_1$) calculates the overlapping rate between the candidate knowledge graph $\mathbf{G}^k_{ti}$ and the full knowledge graph $\mathbf{G}_{\rm goal}$ without considering inference nodes (white and gray nodes in Fig.~\ref{fig:pipeline}). 
We define a match as same entities (nodes)
between two knowledge graph.
Then calculate the story entity overlapping rate by
\begin{equation}
    r_1(\mathbf{G}^k_{ti}, \mathbf{G}_{\rm goal}) = \frac{\sum_{j}\sum_{l}{\mathbb{I}(e_{j, \mathbf{G}^k_{ti}}= e_{l,\mathbf{G}_{\rm goal}})}}{\text{size of } \mathbf{G}_{\rm goal}} 
\end{equation}
where $\mathbb{I}(e_{j, \mathbf{G}^k_{ti}} = e_{l,\mathbf{G}_{\rm goal}}) = 1$ when there is a match between entity $e_{j, \mathbf{G}^k_{ti}} \in \mathbf{G}^k_{ti}$ and entity $e_{l,\mathbf{G}_{\rm goal}} \in \mathbf{G}_{\rm goal}$, otherwise $0$.

We then calculate the overlapping rate between $\widetilde{E^{k}_{ti}}$ and goal knowledge graph with inferences, $\mathbf{G}'_{\rm goal}$ , as \textit{inference overlapping score} $r_2$,
\begin{equation}
    r_2(\widetilde{E^{k}_{ti}}, \mathbf{G}'_{\rm goal}) = \frac{\sum_{j}\sum_{l}{\mathbb{I}(\hat{e}_{j, \widetilde{E^{k}_{ti}}}= e_{l,\mathbf{G}'_{\rm goal}})}}{\text{size of } \mathbf{G}'_{\rm goal}} 
\end{equation}
where $\hat{e}_{j, E^{s_i}_t} \in \widetilde{E^{k}_{ti}}$ and $e_{l,\mathbf{G}'_{\rm goal}} \in \mathbf{G}'_{\rm goal}$. 

\begin{figure}
    \centering
    \includegraphics[width=0.5\textwidth]{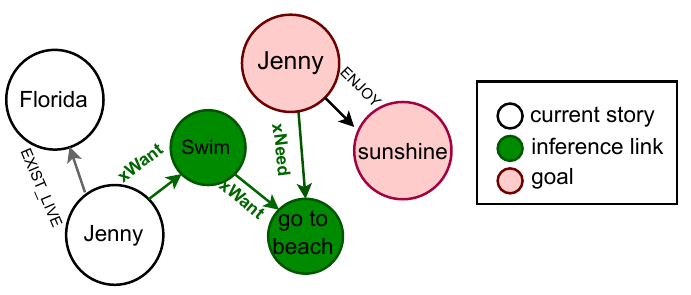}
    \caption{A demonstration of finding the inference link between current story and the goal.}
    \label{fig:link}
\end{figure}

We also consider constructing an inference link between current story and the goal.
When we generate $\widetilde{E^{k}_{ti,{\rm event}}}$ and $\widetilde{E_{\rm goal, event}}$ in Section~\ref{sec:infer}, 
we obtain $\sum_{j=1}^{m_1} {n_2}^{j}$ and $\sum_{j=1}^{m_2} {n_2}^{j}$ inference links, where $n_2$ is the beam size of the output of $\text{COMET}^{20}_{20}$.
For example, when generating $\widetilde{E_{\rm goal, event}}$, the \texttt{xNeed} of the goal ``enjoy sunshine'' is ``go to beach''.
Then we develop a link from goal---``go to beach $\rightarrow$ enjoy sunshine''.
When generating $\widetilde{E^{k}_{ti,{\rm event}}}$, the \texttt{xWant} of the prompt ``live in Florida'' is ``swim'' and the \texttt{xWant} of ``swim'' is ``go to beach''. Then we successfully develop a link between the prompt and the goal---``live in Florida $\rightarrow$ swim $\rightarrow$ go to beach''. An example is shown in Fig.~\ref{fig:link}.

Hence, we calculate the semantic similarity between all events in $\widetilde{E^{k}_{ti,{\rm event}}}$ and $\widetilde{E_{\rm goal, event}}$ by Sentence-BERT~\cite{reimers-2019-sentence-bert}.
When the max semantic similarity score is over threshold\footnote{we use 80\% in this paper.}, we
set $R(\mathbf{G}^k_{ti}) = 1$ and generate stories based on this link. 
We obtain the \textit{topk} continuation knowledge graphs with the \textit{topk} highest graph similarity scores. They will be used to produce continuations further.
We always keep a total of $K$ knowledge graphs (reader models) when generating stories for each index of the sentence in the story. 
Thus the full generation process is implemented as a form of beam search.

\subsection{Continuation Candidate Generation}
\label{sec:gen}
Given \textit{topk} continuation knowledge graph 
$\mathbf{G}^{k}_{ti}$ and its corresponding event $e^k_{ti}$, we generate story continuations.
We consider the conditional sentence generation as a infilling task \cite{taylor1953cloze}. 

\textbf{Templates.}
A set of templates $T_{ti}^k$ are generated on each event $e^k_{ti}$.
For example, one of the templates generated on \texttt{swim} are \texttt{[subject] <mask> <mask> swim <mask>}.
The \texttt{[subject]} of the sentence is (1)~the same subject with the previous sentence, (2)~no fixed (\texttt{<mask>}), or (3)~any characters in previous story history (See Appendix~\ref{app:template}).

\textbf{Text Infilling.}
We fine-tune RoBERTa \cite{liu2019roberta} on story datasets (\S\ref{sec:exp}).
Details are shown in Appendix~\ref{sec:roberta}.
All the templates $T_{ti}^k$ are filled by fine-tuned RoBERTa and we obtain a number of continuation candidates $S^k_{(t+1)i} = \{s^k_{(t+1)i,1}, ... s^k_{(t+1)i,m}\}$ for each event $e^k_{ti}$ where $m$ is the number of templates of each entity. 

\textbf{Filtering.}
We fine-tune GPT-2 \cite{radford2019language} on different datasets(See \S\ref{sec:exp})
and filter the continuation candidates by calculating their conditional probability $\mathbb{P}_{s}$ with it: %
$
    \mathbb{P}_{s} = \prod_{j=1}^{n}\mathbb{P}(\text{X}_j | \text{X}_1, ..., \text{X}_{j-1})
$
,where $n$ is the length of the sentence $s$ and $\text{X}_j$ is the $j$th token in sentence $s$.
We only keep one sentence $s^k_{(t+1)i} \in S^k_{(t+1)i}$ with the highest probability for each event $e^k_{ti}$ 
and append $s^k_{(t+1)i}$ to the $i$th story, where $i \in [1,K]$.

\section{Experiments}
\label{sec:exp}
After proving that the knowledge graph acquisition technique is able to capture the information that natural language story conveys in Appendix~\ref{app:kg_eva}.
We evaluate our system with two experiments, comparing \sysname{} to two strong neural language model story generators on the the dimensions of coherence and controllability.

\textbf{Datasets.}
We conduct the experiments on three datasets:
\begin{itemize}[noitemsep,topsep=0pt,itemsep=0pt,leftmargin=*]
    \item ROCStories\cite{mostafazadeh2016corpus}: contains $98,159$ five-sentence stories involving common-sense scenarios.
    The first and last sentence of one story are used as the prompt and the goal, respectively.
    \item Writing Prompts (WP) \cite{fan2018hierarchical}: a large collection of user-generated stories along with their associated prompts from Reddit.
    Average story length is $59.35$ sentences, which provides longer and complicated stories than ROCStories.
    We extract high-level plots following \citeauthor{ammanabrolu20automated}\shortcite{ammanabrolu20automated}, which are used as prompts and goals to generate stories.
    
    \item Fairy tale stories (FT): following \cite{ammanabrolu2020bringing}, we scraped $695$ stories in fairy tales genre from Wikipedia, which provides a new genre of stories. 
    Average story length is 24.80 sentences.
    Same with Writing Prompts, extracted high-level plots are used as prompts and goals to generate stories.
\end{itemize}

\textbf{Baselines.}
We evaluate our model against two strong baselines:\footnote{Two potential baselines were considered but not pursued. \citet{tambwekar2018controllable} is goal-driven but does not produce natural language without manual intervention. The system by \citet{rashkin2020plotmachines} accepts unordered outline terms but the results of the original paper could not be reproduced at the time of writing.}
\begin{itemize}[noitemsep,topsep=0pt,itemsep=0pt,leftmargin=*]
    
    \item C2PO\cite{ammanabrolu20automated}: 
    uses COMET\cite{bosselut2019comet} to generate successor and predecessor events, performing a bi-directional search from a given start event and a given end event. 
    It uses a subset\footnote{C2PO uses \texttt{xNeed} and \texttt{xWant} relations.} of social interaction relations generated by COMET to build a commonsense link between the prompt and the goal, without constraining language model.
    For fair comparison, following \citeauthor{ammanabrolu20automated}, \sysname{} and C2PO generate a story piece-by-piece given a set of goals.
    Details about this baseline can be found in Appendix~\ref{app:c2po_implement}.
    
    \item \citeauthor{goldfarb2020content}~\shortcite{goldfarb2020content}: 
    trains BART\cite{lewis2020bart} to generate plots on the given prompt and then transform it
    into a story with an ensemble of rescoring model on Writing Prompts dataset. 
    For fair comparison,
    we set \sysname{} and \citeauthor{goldfarb2020content}~\shortcite{goldfarb2020content} up for generating a story piece-by-piece given a set of goals.
    We firstly follow \citeauthor{ammanabrolu20automated}~\shortcite{ammanabrolu20automated}, to extract high-level plots from Writing Prompts.
    We then train the model on the Writing Prompts using these extracted
    plots as
    the prompts,
    and sections in between each of these extracted plot points as the story.
    More details about training can be found in Appendix~\ref{app:cp_implement}.

\end{itemize}

\begin{table*}[!tbh]
    \centering
    \label{TestTable}
    \begin{subtable}{\textwidth}
            \footnotesize
            \centering
            \setlength\tabcolsep{1.8pt} %
            \begin{tabular}{c|c|lll|lll|lll||l|l|l|l}
                \toprule
                \multirow{2}{*}{\textbf{Models}} & \textbf{Data}&
                 \multicolumn{3}{c|}{\textbf{Logical Sense }} & \multicolumn{3}{c|}{\textbf{Enjoyable}} & \multicolumn{3}{c||}{\textbf{Fluency}} & \multicolumn{2}{c|}{\textbf{Self-BLEU-2}$\mathbf{\downarrow}$}&
                 \multicolumn{2}{c}{\textbf{Self-BLEU-3}$\mathbf{\downarrow}$}\\\cline{12-15}
                 
                & \textbf{set}  &Win\%&Lose\%&Tie\%&Win\%&Lose\%&Tie\%&Win\%&Lose\%& Tie\%
                &\sysname{} & Baseline & \sysname{} & Baseline\\

                \midrule
                \sysname{} vs CP
                &WP&
                  \textbf{72.0}**&18.7&9.3&
                  \textbf{52.0}&40.0&8.0&
                  \textbf{74.7}**&18.7&6.7
                  &.137&\textbf{.103}&.098&\textbf{.070}
                   \\
                \midrule
                
                \multirow{2}{*}{\sysname{} vs C2PO}
                &ROC&
                  \textbf{72.9}**&17.1&10.0&
                  \textbf{71.4}**&21.4&7.1&
                  \textbf{62.9}**&24.3&12.9
                  &\textbf{.045}**&.261& \textbf{.035}**&.169 \\
                
                &FT&
                  \textbf{58.7}**&22.7&18.7&
                  \textbf{60.0}**&22.7&17.3&
                  \textbf{53.3}**&21.3&25.3
                  &\textbf{.095}**&.249&\textbf{.072}**&.155 \\
                \bottomrule
            \end{tabular}
        
        \caption{Evaluation results on coherence and diversity.}
        \label{tab:exp_coherence}
    \end{subtable}

    \begin{subtable}{\textwidth}
            \footnotesize
            \centering
            \setlength\tabcolsep{0.11pt} %
            \begin{tabular}{c|c|lll|lll||l|l|l|l|l|l|l|l}
                \toprule
                \multirow{2}{*}{\textbf{Models}} & \textbf{Data}&
                 \multicolumn{3}{c|}{\textbf{Goal}} & \multicolumn{3}{c||}{\textbf{Quality}}&
                 \multicolumn{2}{c|}{\textbf{BLEU-2$\uparrow$}}&
                 \multicolumn{2}{c|}{\textbf{BLEU-3$\uparrow$}}&
                 \multicolumn{2}{c|}{\textbf{ROUGE-L$\uparrow$}}&
                 \multicolumn{2}{c}{\textbf{S-M$\uparrow$}}\\\cline{9-16}
                
                &\textbf{set} & Win\%&Lose\%&Tie\%&Win\%&Lose\%&Tie\%
                &
                \sysname{} & Baseline &\sysname{} & Baseline&\sysname{} & Baseline
                &\sysname{} & Baseline\\

                \midrule
            
                \sysname{} vs CP
                &WP&
                \textbf{57.1}*&31.4&11.4&
                  \textbf{54.3}&38.6&7.1
                  &\textbf{.204}*&.099&\textbf{.169}*&.070&
                  \textbf{.315}*&.244&
                  \textbf{.065}&.046\\
                \midrule
                
                \multirow{2}{*}{\sysname{} vs C2PO}
                &ROC&
                \textbf{73.3}**&16.0&10.7&
                \textbf{74.7}**&20.0&5.3 
                  
                  &.334&\textbf{.361}&.290
                  &\textbf{.329}
                  &\textbf{.428}&.426&
                  \textbf{.121}&.119\\
                
                &FT&
                \textbf{56.0}*&33.3&10.7&
                  \textbf{61.3}**&26.7&12.0
 
                  &\textbf{.110}&.089&\textbf{.079}&.065
                  &\textbf{.198}&.182
                  &.044&\textbf{.058}
                  \\
                \bottomrule
            \end{tabular}
        \caption{Evaluation results on controllability and coverage with respect to gold stories.}
        \label{tab:exp_cotrol}
    \end{subtable}%
\caption{Evaluation results, 
showing the percentage of participants who preferred the first system, second system, or thought the systems were equal. CP indicates \citep{goldfarb2020content}.
Each system is conditioned on the same test-set prompts and same goal. * indicates results are significant at $p<0.05$ confidence level; ** at $p<0.01$ using a Wilcoxan sign test on win-lose pairs. 
See results about majority votes and agreement in Appendix~\ref{app:eva_results}.
}
\label{tab:exp}
\end{table*}

\subsection{Story Coherence Evaluation}
\label{sec:coherence_exp}
We firstly seek to understand whether \sysname{} improves the coherence and quality of the generated story.
To generate the stories, we randomly selected $15$ stories from each dataset and
condition \sysname{} and baselines with same prompts and goals.
More details can be found in Appendix~\ref{app:exp_coherence}.

\textbf{Human Evaluation.}
We first evaluate coherence using human participant evaluation, asking a set of questions that includes dimensions such as logical coherence, enjoyability and fluency. 
Variations of these questions have been used to evaluate other story generation systems (cf. \cite{purdy2018predicting,tambwekar2018controllable,ammanabrolu2020story,ammanabrolu2020automated,Castricato2021Five, peng2021inferring}).
We focus on dimensions involving overall perceptions of narrative coherence:
\begin{itemize}[noitemsep,topsep=0pt,itemsep=0pt,leftmargin=*]
    \item The story's sentences MAKE MORE SENSE given sentences before and after them: evaluates local causality and commonsense reasoning.
    \item The story is more ENJOYABLE: indicates story value.
    \item The story uses more FLUENT language: indicates story readability.
\end{itemize}

\noindent
Each human participant reads a randomly selected subset of story pairs, comprised of one story from \sysname{} and one from baselines.
For the above three questions, participants answered which story best met the criteria. 
Details about human study can be found in Appendix~\ref{app:coherence_setup}.

Table~\ref{tab:exp_coherence} (left) shows the percentage of times stories from each system are preferred for each metric.
\sysname{} improves the perception of narrative coherence of generated narratives, and also produce more enjoyable and fluent stories with the help of common-sense relations.
With the help of transformer-based language model,
\citeauthor{goldfarb2020content}~\shortcite{goldfarb2020content} achieves comparable enjoyability with \sysname{} but fail to guarantee local causality in the story.
C2PO only applies a limited number of ATOMIC relations to conduct a bi-directional search between the prompt and the goal, which cannot guarantee coherence between all the sentences in the story.
At the same time, the generated stories by C2PO are made up of highly constrained, templated text, which also decrease the fluency and enjoyability of the story.

\textbf{Automatic Metrics.}
We also evaluate diversity of generated stories by measuring Self-BLEU scores \cite{zhu2018texygen}.
For each generated story, we take one sentence as the hypothesis and the others as references
and calculate the BLEU score, repeating for every sentence in the story. 
The averaged BLEU score of its generated stories is defined as the self-BLEU score of the model.
A lower self-BLEU score indicates more diversity of the stories. 
Table~\ref{tab:exp_coherence} (right) shows that
\sysname{} significantly outperforms C2PO on the dimension of ``diversity'', because \sysname{} adopts transformer-based language model to generate stories, however, C2PO uses templated and limited range of COMET to generate stories.
Compared with \citeauthor{goldfarb2020content}~\shortcite{goldfarb2020content}, which also applies transformer-based language model, \sysname{} has comparable results on the diversity of generated stories.

\subsection{Controllability Evaluation}
\label{sec:control_exp}

We assess whether \sysname{} is able to achieve the given goal, as measured by human evaluation and automated metrics. 
Generated stories evaluated in Section~\ref{sec:coherence_exp} are reused for evaluating controllability.

\textbf{Human Evaluation.}
Human participant firstly read a  prompt and a goal in natural language.
They then read a randomly selected generated story pairs---one from \sysname{} and one from our baselines.
They then answered which one better met the criteria (See Appendix~\ref{app:controllability_setup}):
\begin{itemize}[noitemsep,topsep=0pt,itemsep=0pt,leftmargin=*]
    \item Which story better FOLLOWS A SINGLE TOPIC TO ACHIEVE GOAL: evaluates perceptions of global coherence for the entire story.
    \item Which story has higher QUALITY to achieve the goal: measures overall perceived story quality.
\end{itemize}
Table~\ref{tab:exp_cotrol} (left) shows that, 
given the same goal and prompt, \sysname{} performs significantly better than C2PO on producing more goal-directed and higher-quality stories with fair agreement (See Appendix~\ref{app:eva_results}).
C2PO restricts its bi-directional search from a given start event and an end event in a subset of ATOMIC relations generated by COMET.
Hence, it often fails to find a coherent commonsense link to develop a story,
which makes human participants hard to understand (see Table~\ref{tab:examples}).

\sysname{} also significantly outperforms \citeauthor{goldfarb2020content}~\shortcite{goldfarb2020content} on the dimension of ``Goal'', with fair agreement.
Compared to \citeauthor{goldfarb2020content}~\shortcite{goldfarb2020content}---which applies transformer-based language model to learn how to achieve a goal,
\sysname{} successfully produce a much more goal-directed story with the help of external common-sense knowledge.
On the dimension of ``quality'', \sysname{} is preferred but the result is not statistically significant when ties are considered, 
which indicates \sysname{} improves controllability while retaining high quality of stories.

\textbf{Automatic Metrics.}
We also evaluate controllability by measuring the the following three metrics with respect to the gold stories.
\begin{itemize}[noitemsep,topsep=2pt,itemsep=2pt,leftmargin=*]
    \item \textit{Sentence mover’s similarity (S-M)} \cite{clark2019sentence}: 
    Evaluate stories in a continuous space using word and sentence embeddings. Larger sentence mover’s similarity indicates higher similarity between generation and gold story.
    \item \textit{BLEU scores}\cite{papineni2002bleu}: 2-gram, 3-gram BLEU scores are reported.
    \item ROUGE-L \cite{lin2004rouge} scores: Higher score indicates better coverage.
\end{itemize}

Right side of Table~\ref{tab:exp_cotrol} shows that
\sysname{} significantly outperforms \citeauthor{goldfarb2020content}~\shortcite{goldfarb2020content} in the coverage with respect to gold stories in Writing Prompts dataset.
It justifies our human evaluation results that \sysname{} generates much more goal-directed stories.
Compared with C2PO on ROCStories and fairy tale stories, 
\sysname{} achieved the comparable results.
The story generation is guided by goal events, so it is likely to produce very different story from gold stories, but with same goal (see Table~\ref{tab:examples}).

\section{Conclusions}

Neural language models are widely used to produce text, including stories. 
However, they struggle with maintaining \textit{story coherence}---the logical progression of events---and goal-directedness.
Our framework---\textit{Story Generation with Reader Models} (\sysname{})--- augments neural language models with a reader model.
This reader model---an explicit knowledge graph---approximates the reader’s beliefs about the story world.
\sysname{}  
increases the story coherence by 
producing continuations that directly reference entities in this reader model.
Goal-directedness is achieved by choosing continuations that add desired entities to the reader's inferred set of beliefs.
A thorough experimental study shows that \sysname{} produces significantly more coherent and goal-directed stories than two strong baselines on three datasets.

\section{Broader Impact}
Our system faces the same potential pitfalls as other contemporary language learning systems.
It is prone to echoing the biases present in the dataset \cite{sheng2019woman} and generate non-normative text (i.e. in violation of social norms).
No existing automated storytelling systems is able to entirely eliminate these biases, though stories can be used to teach language models to reduce non-normative continuations~\cite{Peng2020ReducingNT}.
Fictional stories that are presented to readers as non-fictional can be used to influence~\cite{green00} or misinform.
Future work may enable real-world facts to be injected into the knowledge graph of a similar system for the purposes of journalism or misinformation.
However, because our graph expansion method relies on ConceptNet5 \cite{speer2013conceptnet} and $\text{COMET}^{20}_{20}$~\cite{Hwang2021COMETATOMIC2O} for inference, our system is prone to process and produce simple stories.

The ability to produce coherent and goal-directed stories has downstream applications beyond automated story-telling.
In particular, this is the first work that increases generative coherence and control by reasoning about the changes to the knowledge graph.

\bibliography{anthology,custom}
\bibliographystyle{acl_natbib}
\clearpage
\appendix
\section{Graph Expansion Details}
\label{app:expand}
For current knowledge graph $\mathbf{G}_{ti} \in \mathbf{G}_{t}$, where $i \in [1, K]$, we expand each concept by $m_1$ depth. 
We firstly expand all the physical entities in current knowledge graph $\mathbf{G}_{ti}$ :
\begin{enumerate}[noitemsep,topsep=0pt,itemsep=0pt,leftmargin=*]
    \item For every physical entity in $\mathbf{G}_{ti}$, we infer as set of relevant entities to construct $E_{ti,{\rm entity}}$ from ConceptNet.
    \item For each physical entity in $E_{ti, {\rm entity}}$  (i.e. $k_1$th event, $e^{k_1}_{ti,{\rm entity}}$, where $k_1 \in [1,n_1]$), keep inferring relevant entities to construct $E^{{k_1},2}_{ti,{\rm entity}}$ until we get $m_1$ deep inference, $E^{{k_1},m_1}_{ti,{\rm entity}}$.
    \item Hence, we obtain an inference event set $\widetilde{E^{k_1}_{ti,{\rm entity}}} = \{e^1_{ti,{\rm entity}}\}$ $\cup E^{k_1,2}_{ti,{\rm entity}}$ $\cup$ $...$ $\cup$ $E^{k_1,m_1}_{ti,{\rm entity}}$ for entity $e_{ti,{\rm entity}}^{k_1}$.
\end{enumerate}

Next, we consider social events:
\begin{enumerate}[noitemsep,topsep=3pt,itemsep=3pt,leftmargin=*]
    \item Construct another inference event set $E_{ti,event}$ by inferring social interaction of current story history by $\text{COMET}^{20}_{20}$. $n_2$ is the beam size of the output of $\text{COMET}^{20}_{20}$.
    For example, the \texttt{xEffect} of ``graduate from college'' is ``get degree'' (green nodes in red bubbles of Fig.~\ref{fig:pipeline}).
    \item For each event in $E_{ti,event}$ (i.e. $k_2$th event, $e^{k_2}_{ti,event}$), keep inferring social interaction to construct $E^{k_2,2}_{ti,event}$ until we get $m_1$ level inference, $E^{k_2,m_1}_{ti,event}$.
    Hence, we obtain inference event set $\hat{E}^{k_2}_{ti,event}$ $= \{e^1_{ti,event}\}$ $\cup E^{k_2,2}_{ti}$ $\cup ... \cup$ $E^{k_2,m_1}_{ti,event}$ for $e_{ti,event}^{k_2}$.
    \item For each event in new inference event set---$\hat{E}^{k_2}_{ti,event}$, we infer all the relevant entities by $\text{COMET}^{20}_{20}$ and ConceptNet to construct $\tilde{E}^{k_2}_{ti,event}$.
    For example, ``get degree'' is relevant with ``job''.
    Then we obtain inference set $ \widetilde{E^{k_2}_{ti,event}} =$ $\hat{E}^{k_2}_{ti,event} \cup \tilde{E}^{k_2}_{ti,event}$ for event $e^{k_2}_{ti,event}$.
\end{enumerate}

Hence, for $\mathbf{G}_{ti}$, we totally obtain $n$ entity candidates $e_{ti}^k$ $\in$ $\{E_{ti,{\rm entity}}$ $\cup$ $E_{ti,event}\}$, where $k \in [1,n]$ and $n = n_1 + n_2$ and its corresponding
knowledge graph candidates $\mathbf{G}'_{ti} = \{\mathbf{G}^{'1}_{ti}, ..., \mathbf{G}^{'n}_{ti}\}$, where $\mathbf{G}^{'k}_{ti} = \mathbf{G}_{ti} \cup \widetilde{E^{k}_{ti}}$ , where $ \widetilde{E^{k}_{ti}}$ $\subset$ $\{\widetilde{E^{k_1}_{ti,{\rm entity}}}$ $\cup$ $\widetilde{E^{k_2}_{ti,event}}\}$.
Totally, we obtain $\mathbf{G}'_t$ $=$ $\{\mathbf{G}'_{t1},$ $...,$ $\mathbf{G}'_{tK}\}$.

We also expand the depth of the graph for goals by $m_2$ via similar means:
\begin{enumerate}[noitemsep,topsep=0pt,itemsep=0pt,leftmargin=*]
    \item For each node in $\mathbf{G}_{\rm goal}$, we expand all the relevant entities with Conceptnet by $m_2$ depth to construct $\widetilde{E_{\rm goal, enitity}}$.

    \item Construct event inference set $E_{\rm goal}^1$ by inferring social interaction of story goal by $\text{COMET}^{20}_{20}$.
    For example, the \texttt{xNeed} of ``enjoy sunshine'' is ``go to beach'' (green nodes in Fig.~\ref{fig:pipeline}).
    
    \item Keep inferring social interaction of all the events in $E^{n}_{\rm goal}$ to construct $E^{n+1}_{\rm goal}$ until we get $E^{m_2}_{\rm goal}$.
    $\hat{E}_{\rm goal} = \{E^{1}_{\rm goal}$,..., $E^{m_2}_{\rm goal}\}$. 
    
    \item For each event in $\hat{E}_{\rm goal}$, and all the nodes in $\mathbf{G}_{\rm goal}$, we infer all the relevant entities to construct $\tilde{E}_{\rm goal}$ by $\text{COMET}^{20}_{20}$ and ConceptNet.
    $\widetilde{E_{\rm goal,event}} = \hat{E}_{\rm goal} \cup \tilde{E}_{\rm goal}$.

    \item Obtain the updated knowledge graph $\mathbf{G}'_{goal} = \mathbf{G}_{\rm goal} \cup \widetilde{E_{\rm goal, event}} \cup \widetilde{E_{\rm goal, enitity}}$.
    
\end{enumerate}

\section{Implementation Details}

\subsection{Semantic Role Labeling Using VerbAtlas}
\label{sec:verbatlas-srl}
The SRL model provides the automatic identification and labeling of argument structures of stories.
For example, it extracts \texttt{\text{`verbatlas'}: \text{`EXIST\_LIVE'},  \text{`args\_words'}: \{\text{`Theme'}: \text{`Jenny'}, \text{`Attribute'}: \text{`Florida'}\}} from ``\textit{Jenny lived in Florida}''.
Verbs in the story will be represented as the VerbAtlas frame.
For example, \texttt{``live''} is represented as \texttt{``EXIST\_LIVE''}.

For the semantic role labeling model (SRL), we use a fine-tuned transformer model proposed by \cite{shi2019simple} which is the current state-of-the-art for English SRL. It is a BERT \cite{devlin2019bert} model with a linear classification layer trained on the Ontonotes 5.0 dataset to predict PropBank SRL.  
We use an open-source implementation \footnote{\href{https://github.com/Riccorl/transformer-srl}{https://github.com/Riccorl/transformer-srl}}, which is based on the official AllenNLP BERT-SRL model \footnote{\href{https://demo.allennlp.org/semantic-role-labeling}{https://demo.allennlp.org/semantic-role-labeling}}. Trained with the following hyperparameters:
\begin{compactitem}
    \item Batch size: 32
    \item Dropout for the input embeddings: 0.1
    \item Learning rate: $5e^{-5}$
    \item Optimizer: Adam
    \item Total Epochs: 15
\end{compactitem}

Then, we use the mappings from Propbank frames to VerbAtlas \cite{di2019verbatlas} classes to return the correct corresponding VerbAtlas classes instead of Propbank's \cite{palmer2005proposition}. The direct mapping is possible because, for every VerbAtlas class, there is only one ProbBank frame, which allows us to utilize the rich content provided by VerbAtlas while using the same model initially trained to predict ProbBank. 

\subsection{Common-sense Inference Types}
\label{app:comet_inference}
Table~\ref{tab:def} shows the inference relations of ATOMIC we use for inferring physical entity.
Table~\ref{tab:def-2} shows the inference relations of ATOMIC we use for inferring events.

\begin{table}
\footnotesize
\centering
\setlength\tabcolsep{1pt} %
\begin{tabular}{l|l}
\toprule
 Type & Definition\\
\midrule
\texttt{AtLocation}&	located or found at/in/on\\
\texttt{CapableOf} &	is/are capable of\\
\texttt{HasA} &	has, possesses or contains\\
\texttt{HasProperty}&	can be characterized by being/having\\
\texttt{MadeOf}&	is made of\\
\texttt{MadeUpOf}&	made (up) of\\
\texttt{{}MotivatedByGoal}&	is a step towards accomplishing the goal\\
\texttt{UsedFor} &	used for\\
\texttt{PartOf} &	is a part of\\
\bottomrule
\end{tabular}
\caption{Definitions of the selected types of ATOMIC relations that we used for graph expansion.}
\label{tab:def}
\end{table}
\begin{table}
\footnotesize
\centering
\setlength\tabcolsep{1pt} %
\begin{tabular}{l|l}
\toprule
 Type & Definition\\
\midrule
\texttt{xWant} & as a result, \texttt{PersonX} wants \\
\texttt{xNeed} & before this event, \texttt{PersonX} needed\\
\texttt{xEffect} & as a result, \texttt{PersonX} will\\
\texttt{oWant} & as a result, \texttt{PersonY} or others want\\
\texttt{oEffect} & as a result, \texttt{PersonY} or others will\\
\bottomrule
\end{tabular}
\caption{Definitions of the selected types of ATOMIC relations that we used for graph expansion.}
\label{tab:def-2}
\end{table}

\subsection{Continuation Candidate Generation Details}
\label{app:template}
We first use pattern package\footnote{ \href{https://github.com/clips/pattern}{https://github.com/clips/pattern}} to extract tense of prompt and run verb conjugation for each event $e^k_{ti}$.
Then we use nltk toolkit\footnote{ \href{https://www.nltk.org/}{https://www.nltk.org/}} to extract adjective and noun set from  $e^k_{ti}$.
Each event $e^k_{ti}$ is represented as three sets---verb, noun and adjective.
For example, when prompt is ``Jenny celebrated her birthday.'', the event ``get a gift'' is split into three sets $\{\{``got''\}, \{gift\}, \{\}\}$.
When prompt is ``Jenny eats a lot of ice cream.'', event ``get fat'' is split into three sets $\{\{``gets''\}, \{\}, \{fat\}\}$.
We generate templates with the following rules,
\begin{compactitem}
    \item Before the subject token, we put $0 \sim 5$ \texttt{$\langle mask \rangle$} tokens.
    \item Subject tokens are (1) Previous occurred subjects, i.e. \textit{``Jenny'', ``Bob''}; (2) \texttt{$\langle mask \rangle$}.
    \item Between the subject token and the verb token, we put $0 \sim 2$ \texttt{$\langle mask \rangle$}.
    \item Between the verb token and the adjective token, we put $0 \sim 2$ \texttt{$\langle mask \rangle$}.
     \item Between the adjective token and the noun token, we put $0 \sim 2$ \texttt{$\langle mask \rangle$}.
    \item After the noun token, we put $0 \sim 8$ \texttt{$\langle mask \rangle$} tokens.
\end{compactitem}
Examples (we show three) are as follows when $e^k_{ti} = $\texttt{``go to beach''} and prompt is ``Jenny lived in Florida.'',
\begin{compactitem}
    \item \texttt{\texttt{$\langle mask \rangle$} Jenny went \texttt{$\langle mask \rangle$} beach \texttt{$\langle mask \rangle$}}.
    \item \texttt{Jenny \texttt{$\langle mask \rangle$} went \texttt{$\langle mask \rangle$} beach}.
    \item \texttt{Jenny \texttt{$\langle mask \rangle$} \texttt{$\langle mask \rangle$} went beach \texttt{$\langle mask \rangle$} \texttt{$\langle mask \rangle$} \texttt{$\langle mask \rangle$}}.
\end{compactitem}

\subsection{RoBERTa Fine-tuning}
\label{sec:roberta}

We fine-tune RoBERTa-large~\cite{liu2019roberta} on ROCStories~\cite{mostafazadeh2016corpus}, Writing Prompts\cite{fan2018hierarchical}, Fairytale stories\cite{ammanabrolu2020bringing} separately to infill the mask tokens in the given text template.
We pre-process the datasets by masking 15\% of all the tokens randomly, concatenating all texts together, and splitting them into chunks of the same length (equal to 128). Each chunk is then used as one training sample.

During fine-tuning, we use the AdamW optimizer~\cite{loshchilov2017decoupled} to train the RoBERTa for 3 epochs with batch size = 8. Other optimizer-related hyperparameters are attached as follows.
\begin{compactitem}
    \item learning rate: $\gamma = 2\times 10^{-5}$
    \item betas: $\beta_1 = 0.9$, $\beta_2 = 0.999$
    \item epsilon: $\epsilon = 10^{-8}$
    \item weight decay: $\lambda = 0.01$
\end{compactitem}

\subsection{Baselines---C2PO\cite{ammanabrolu20automated}}
\label{app:c2po_implement}
We replicate the C2PO model by \citeauthor{ammanabrolu20automated}  using the code published on the paper's public repository. \footnote{\href{https://github.com/rajammanabrolu/C2PO}{https://github.com/rajammanabrolu/C2PO}}
All the encoder and model checkpoints are provided by the author.

\subsection{Baselines--- \citeauthor{goldfarb2020content}~\shortcite{goldfarb2020content}}
\label{app:cp_implement}
We firstly extract high-level outlines with the help of codes provided by \citeauthor{ammanabrolu20automated}~\shortcite{ammanabrolu20automated}\footnote{\href{https://github.com/rajammanabrolu/C2PO/tree/master/Plot-Extraction}{https://github.com/rajammanabrolu/C2PO/tree/master/Plot-Extraction}}.
Then we split the long Writing Prompt stories into sections according to extracted plots.
Each high-level plot is used as prompt and section in between as story.
We fine-tune BART model using the code and parameters published on the paper's public repository. \footnote{\href{https://github.com/PlusLabNLP/story-gen-BART}{https://github.com/PlusLabNLP/story-gen-BART}}

\clearpage
\section{Knowledge Graph Acquisition Evaluation}
\label{app:kg_eva}
We assess whether knowledge graph can acquire the story world state accurately and comprehensively. 
We randomly select $125$ sentences from ROCStories and convert them into knowledge graph triples.
Human participants were asked to validate each graph triples given the sentence and then write down the missing information. 
For example, they need to check whether $\langle jenny, LIKE, beach \rangle$ given ``\textit{Jenny likes beach and sunshine}'' is correct and write down the missing concept, ``\texttt{sunshine}''. 
The detail of this study is shown in Appendix~ \ref{app:set-up-kg}.

\begin{table}
\footnotesize
\centering
\begin{tabular}{c|c|c}
\toprule
\textbf{Precision \%}  & \textbf{Recall \%}  & \textbf{\# of triplets} \\
\midrule
$81.96^\ddagger$ & $72.89^\ddagger$ & 255 \\
\bottomrule
\end{tabular}
\caption{Results of evaluating knowledge graph triplets. $\ddagger$
indicates $\kappa$ > 0.4 or moderate
agreement.
}
\label{tab:kg}
\end{table}
Table~\ref{tab:kg} shows the accuracy (precision) and sensitivity (recall) of the extracted knowledge graph triples.
We treat the majority vote from human participants as the ground-truth.
\textit{Precision} is the fraction of extracted triples that are correct rated by human participants. 
\textit{Recall} is the fraction of the triples that are successfully extracted from stories.
Precision, $81.96\%$, shows that the knowledge graph can represent the information in sentences accurately. 
Recall, $72.89\%$, proved that the knowledge graph can represent most of the information in sentences.
Both of these two metrics have moderate agreement.
This indicates that the knowledge graph extracted from sentences matches reader expectations and can be used as story world state upon which to base further story generation.

\clearpage
\section{Experiments Details}
\label{app:exp_coherence}

\paragraph{ROCStories.}
We firstly compare our model with C2PO on ROCStories.
Since the length of ROCStories is 5 sentences.
We randomly select $15$ stories and use the first sentence as prompt and the last sentence as goal to generate stories.

\sysname{} firstly convert first and last sentence as story KG and goal KG.
Then we look ahead $2$ step ($m_1 = m_2 = 2$ in Section~\ref{sec:infer}) and set the max story length as $4$.
When \sysname{} finds a inference link to reach the goal or the graph similarity is over 80\%, the system will \textbf{early stop}. 
Beam size of COMET is set as $5$.
Hence the story length of \sysname{} is $2 \sim 6$ (prompt and goal sentences are counted).

C2PO finds a inference link between the first sentence and the last sentence by generate $3$ event candidates twice going forward and backward, respectively. 
Hence the story length of \sysname{} is $2 \sim 6$ (prompt and goal sentences are counted). 
Examples can be found in Table~\ref{tab:app_roc_ex}.
\begin{table}[!tbh]
\centering
\footnotesize
\setlength\tabcolsep{6pt} %
\begin{tabular}{l}
\toprule
\textbf{\sysname:}\\
\textbf{Justin decided to make dinner for his boyfriend.}\\
So he tried to cook spaghetti.\\
But, he failed.\\
In fact, he ate a lot of spaghetti everyday.\\ So he \textbf{ordered pizza} instead.\\
\midrule
\textbf{C2PO:}  \\
\textbf{Justin decided to make dinner for his boyfriend.}\\
Justin tries to cook.\\
Justin wants clean up.\\
Justin tries to work.\\
Justin starts to have money.\\
\textbf{Justin ordered pizza.}\\
\midrule
\midrule
\textbf{\sysname:}\\
\textbf{Robin was afraid of flying.}\\
The first time he flew, it scared him.\\
After that he got more nervous. \\
Now all he wanted was to calm his down.\\
He took a deep breath.\\
Robin \textbf{felt less nervous} now.\\
\midrule
\textbf{C2PO:} \\
\textbf{Robin was afraid of flying.}\\
Robin wants to save himself.\\
Robin wants to find a job.\\
Robin tries to have a job.\\
Robin wants to be nervous.\\
\textbf{Robin felt a lot less nervous about her flight.}\\
\midrule
\midrule
\textbf{\sysname:}\\
\textbf{Anna was having a bad day at work.}\\
She wished she had stayed home.\\
She was glad that friends invited her to movie night.\\
She had a fun time with them.\\
She is \textbf{grateful and pleased} now.\\
\midrule
\textbf{C2PO:} \\
\textbf{Anna was having a bad day at work.}\\ 
Anna tries to get some rest. \\
Anna wants to get up. \\
Anna begins go to the store.\\
Anna starts to have received a gift.\\
\textbf{Anna was grateful and pleased.}\\

\midrule
\midrule
\textbf{\sysname:}\\
\textbf{David noticed he had put on a lot of weight recently.} \\
He starts to eat less. \\
So he exercised more every day.\\ 
And he \textbf{felt much better} afterwards.\\
\midrule
\textbf{C2PO:} \\
\textbf{David noticed he had put on a lot of weight recently.}\\
David starts to eat less.\\
David begins to exercise.\\
David begins to rest.\\
\textbf{He started to feel much better.}\\
\bottomrule
\end{tabular}
\caption{Story examples generated by \sysname{} and C2PO on ROCStories. Bolded words are prompts and goals. The goal of \sysname{} is knowledge graph, so it will generated the goal sentence by itself if \sysname finds a inference link to reach the goal or similarity with goal knowledge graph is over threshold. C2PO will use given goal sentence in the generated stories directly.}
\label{tab:app_roc_ex}
\end{table}

\paragraph{Writing Prompts}
\begin{table}[!tbh]
\centering
\footnotesize
\setlength\tabcolsep{6pt} %
\begin{tabular}{p{0.9\linewidth}}
\toprule
\textbf{\sysname:}\\
\textbf{Sherlock sat in the downstairs sitting room.} When he talked to people in the room. he never got a response back. So he thought about what was wrong. \textbf{Sherlock himself stared into space , deep in thought .} In fact, he never thought of anyone. He thought that he was alone in the world. Then, he found his mate. There, he got a hug from his wife. He asked if he could put his coat on her. \textbf{He straightened up and donned his favourite coat.}\\
\midrule
\textbf{\citeauthor{goldfarb2020content}~\shortcite{goldfarb2020content}}  \\
\textbf{Sherlock sat in the downstairs sitting room.} The room was silent, save for a man sitting in his own chair. the chair made of wood, built of oak. a chair that would make a good bed and a bed that would be comfortable. Looking around at the room, the woman saw a small smile cross Holmes ' face, but she did n't say anything. 
\textbf{Sherlock himself stared into space, deep in thought.} He was cold. The cold. He felt the cold of the night air around him. His eyes looked into the darkness that seemed so much more than the cold and he sat there. The cold took his in. It was cold, and it seemed like forever as his eyes stared into the dark that seemed to be the end. \textbf{He straightened up and donned his favourite coat.}\\
\midrule
\midrule
\textbf{\sysname:}\\
\textbf{Sarah sat with her dog.} she had a companion. her dog. One day she took the animal out with her. They enjoyed the sunshine and weekend. She whispered soothing words into its ear.\textbf{ She tried talking to the dog while she pet it.} Suddenly, she was bited by the dog. She then yelled at the dog. She wanted to get the dog to stop. She finally got the dog to stop.\textbf{ The dog made cute noises.}\\
\midrule
\textbf{\citeauthor{goldfarb2020content}~\shortcite{goldfarb2020content}} \\
\textbf{Sarah sat with her dog}. The girl was young and pretty, but she was also scared. Her mother stood before her, and the woman seemed like hers to her. “ Sarah. ” The girl looked back and forth between them, and the woman, and then his eyes looked at him for a moment, and all she could see was fear, and sadness. Sarah looked away. The man, who was older than the two of herself, and had a dark hairline and dark eyes.
\textbf{She tried talking to the dog while she pet it.} My mind raced. It was the same as it always had. I was the only one in the room and he knew what he wanted, a man who wanted my love, my life. A man who was so perfect for me. The man came to mine in our bedroom, we would go out for a drink. \textbf{The dog made cute noises.}\\
\bottomrule
\end{tabular}
\caption{Story examples generated by \sysname{} and C2PO. Bolded words are prompts and goals.}
\label{tab:app_wp_ex}
\end{table}
We count the length of Writing Prompts by splitting stories with period, question mark and exclamation mark.
To use \sysname{} to generate stories on Writing Prompts,
we first follow \citeauthor{ammanabrolu20automated}~\shortcite{ammanabrolu20automated} to extract high-level plots from Writing Prompts stories.
Then we use the first event in the extracted plots as prompt to start \sysname{} and \citeauthor{goldfarb2020content}~\shortcite{goldfarb2020content}. 
Goal is the second event of extracted plots.\footnote{Goals are only provided to \sysname, because \citeauthor{goldfarb2020content}~\shortcite{goldfarb2020content} is unconstrained story generator.}
We look ahead $2$ step ($m_1 = m_2 = 2$ in Section~\ref{sec:infer}) and set the max story length as $5$.
When \sysname{} finds a inference link to reach the goal or the graph similarity is over 80\%, or $5$ sentences have been generated, the system will \textbf{early stop}. 
Beam size of COMET is set as $5$.
Hence the local story length of \sysname{} is $2 \sim 6$(prompt and goal sentences are counted).
We then append the goal to the story history, feed the third plot as goal and keep generation with the same process until we use all the plots up.

To seed the \citeauthor{goldfarb2020content}~\shortcite{goldfarb2020content}, 
we use the same high-level plots as prompts.
Then we truncate the generated stories to $6$ sentences and append the next plot to story history.
We then keep generating story on the next plot until we use all the plots up.
Totally we randomly select $15$ generated stories\footnote{Each story is generated on one plot and its next plot is used as the goal.} for human evaluation. 
More examples can be found in Table~\ref{tab:app_wp_ex}.

\paragraph{Fairytale stories}
\begin{table}[!tbh]
\centering
\footnotesize
\setlength\tabcolsep{6pt} %
\begin{tabular}{p{0.9\linewidth}}
\toprule
\textbf{\sysname:}\\
\textbf{Bearskin clip his nails.} When he just got a haircut , his hair looks great. He thought he \textbf{looked good} in it. In fact , he has won every competition in the years. In fact , he has got every trophy in the years. However , he never showed it. In fact , he always hid from everyone. When he met other people , he was shy. \textbf{Bearskin dropped his half of ring.}Then , he picked the ring it up. Then , he proposed to her with the ring. \textbf{He was her bridegroom.}
\\
\midrule
\textbf{C2PO:}  \\
\textbf{Bearskin clip his nails.} Bearskin tries to clean them. Bearskin begins wash hands. Bearskin tries to go to school. Bearskin starts to train consistently.
\textbf{Bearskin is good.} Bearskin wants to win. Bearskin wants to celebrate. Bearskin wants to get a ring. \textbf{Bearskin dropped his half of ring.}
Bearskin begins to get a new ring.Bearskin tries buy a ring. Bearskin wants to have a wedding. \textbf{He was her bridegroom.}\\

\midrule
\midrule
\textbf{\sysname:}\\
\textbf{Girl did hard work.} When she was promoted to the new job. \textbf{Girl cried.} She met with a boy. She dated with him for one year. \textbf{Girl got a ring from him.} Then , she got to kiss the boy. Finally , she had sex with the boy. \textbf{She had a baby.}\\
\midrule
\textbf{C2PO:} \\
\textbf{Girl did hard work.} Girl tries to celebrate. Girl tries to go to bed. Girl wants to fall. Girl begins to get hurt.\textbf{ Girl cried to God.} Girl tries to feel better. Girl starts to go to the bathroom. Girl wants to get ready. Girl begins to go to the ring.
\textbf{Girl got a ring.} Girl begins to buy it. Girl wants to open it. Girl tries to open the door. Girl starts to bring the baby to the car. \textbf{She had her baby.}\\
\midrule
\textbf{\sysname:}\\
\textbf{Old soldier returned from war.} Then he found his family. Then he told his family what happened. He said he was yelled at. Then he got very angry and hurt others. Then he was arrested. When he went to court , \textbf{he met the the woman} in black. He met with woman. he never had a relationship with the woman. He began to date with her. He could not move his eyes away from her. \textbf{He observe king 's daughters.}\\
\midrule
\textbf{C2PO:} \\

\textbf{Old soldier returned from war.} Old soldier wants to get a drink. Old soldier starts to go out. Old soldier starts to meet the woman. \textbf{He met with woman.} Old soldier wants to get to know them. Old soldier wants to ask them out. Old soldier starts to get married. Old soldier begins to get married. \textbf{He observe king 's daughters.}\\
\bottomrule
\end{tabular}
\caption{Story examples generated by \sysname{} and C2PO on Fairytale Stories. Bolded words are prompts and goals. The goal of \sysname{} is knowledge graph, so it will generated the goal sentence by itself if \sysname finds a inference link to reach the goal or similarity with goal knowledge graph is over threshold. C2PO will use given goal sentence in the generated stories directly.}
\label{tab:app_ft_ex}
\end{table}

We first follow \citeauthor{ammanabrolu20automated}~\shortcite{ammanabrolu20automated} to extract high-level plots from fairytale stories.
We use the first plot in the extracted plots as prompt to start \sysname{} and C2PO. 
Goal is the second plot of extracted plots.
\sysname{} firstly convert first plot and second plot as story KG and goal KG.
Then we look ahead $2$ step ($m_1 = m_2 = 2$ in Section~\ref{sec:infer}) and set the max story length as $4$.
When \sysname{} finds a inference link to reach the goal or the graph similarity is over 80\%, the system will \textbf{early stop}. 
Beam size of COMET is set as $5$.
Hence the local story length of \sysname{} is $2 \sim 6$ (prompt and goal sentences are counted).
\sysname{} keeps generating story on the previous story and set goal as the next plot of extracted plots until we use all the plots up.

We follow the original paper, we ask C2PO to fill in the section between high-level plots with the inference link it finds as generated stories. 
Totally we randomly select $15$ generated stories\footnote{Each story is generated on one plot and its next plot is used as the goal.} for human evaluation. 
More examples can be found in Table~\ref{tab:app_ft_ex}.

\clearpage
\section{Human Evaluation Details}

\subsection{Knowledge Graph Acquisition Evaluation Set-up}
\label{app:set-up-kg}
\begin{figure}[H]
    \centering
    \includegraphics[width=\columnwidth]{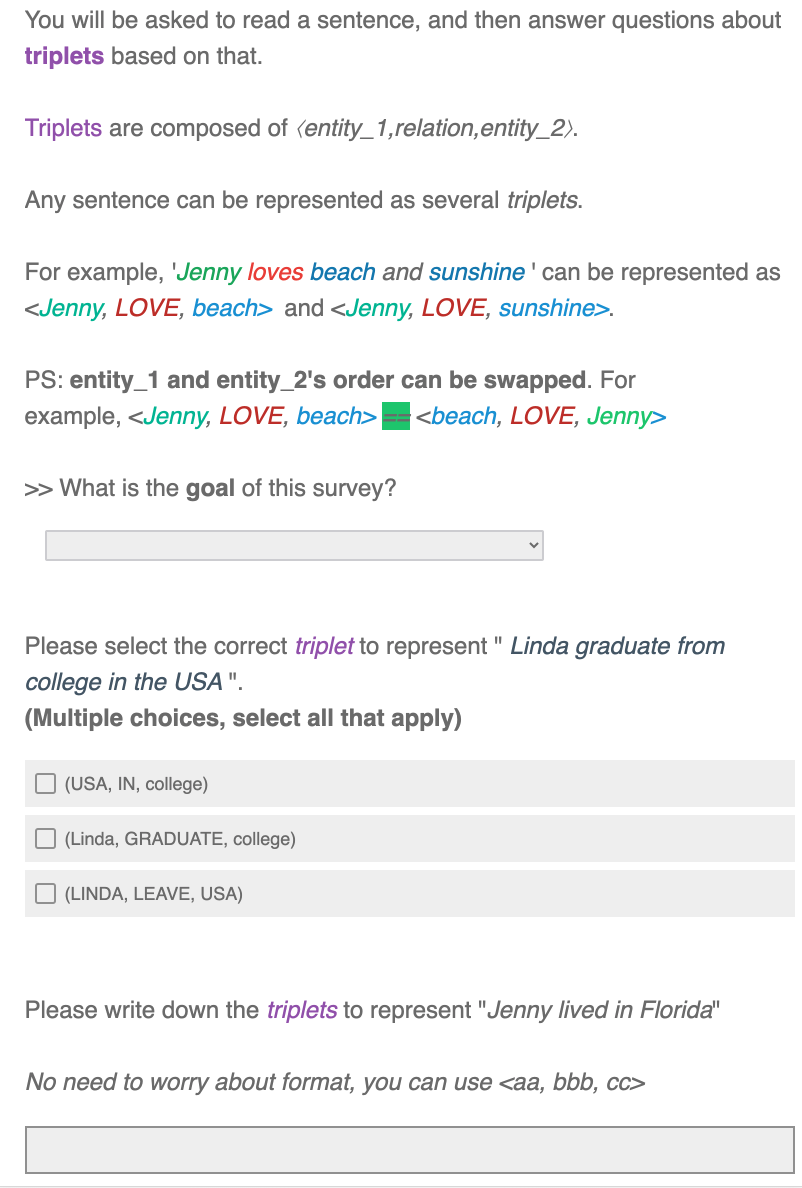}
    \caption{Screenshot of the human study instruction.}
    \label{fig:screenshot-start}
\end{figure}

We ask participants a set of screen questions to make sure they understand our task. 
The details can be found in Figure~\ref{fig:screenshot-start}.
We conduct our studies using the Cloud Research crowdsourcing platform to interact with Amazon Mechanical Turk \cite{litman2017turkprime}. 
Obtaining at least a bachelor's degree and English as their native language are required to take this study.
Participants are required to pass screening questions and then explain their preferences  of each choice in this study with more than $50$ characters,
which helps filter out low-quality responses and ensures the validity of the study. 
Our study was approved by our Institutional Review Board, and we payed participants the equivalent of $\$15$/hr.

\begin{figure}[H]
    \centering
    \includegraphics[width=\columnwidth]{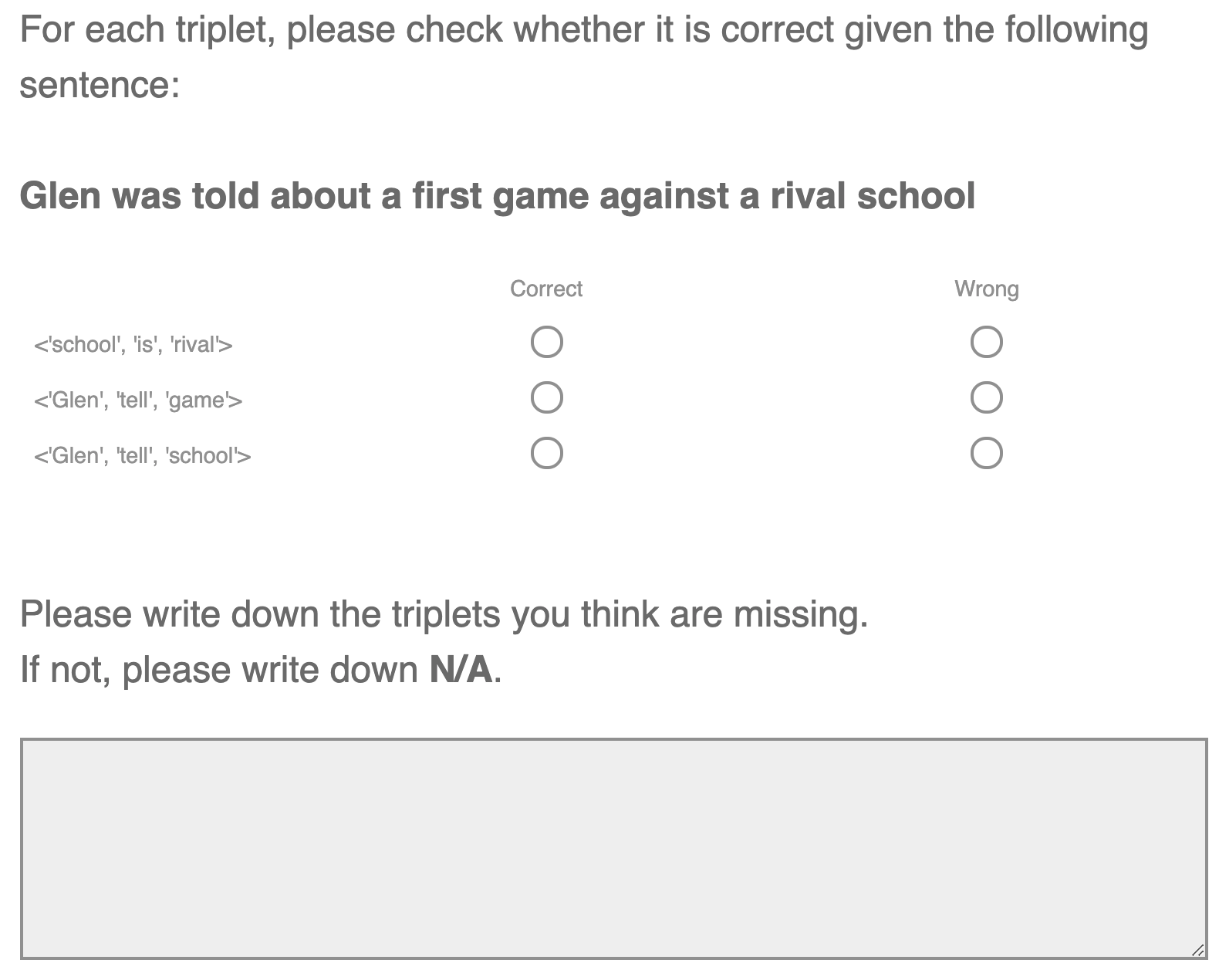}
    \caption{Screenshot of Knowledge Graph Acquisition evaluation.}
    \label{fig:kg_human}
\end{figure}

We assess whether knowledge graph can acquire the story world state accurately and comprehensively. 
We randomly select $125$ sentences from ROCStories and convert them into knowledge graph triplets.
We recruited $30$ participants on a crowdsourcing platform.
Each participant read a randomly selected subset of knowledge graph triplets ($20$ sentences per participant).
They were asked to validate each graph triplets given the sentence and then write down the missing information. 
An example is shown in Figure~\ref{fig:kg_human}.
At least $3$ crowd workers validate each triple and we take the majority vote as the result.

\subsection{Story Coherence Evaluation Set-up}
\label{app:coherence_setup}
\begin{figure}[!tbh]
    \centering
    \includegraphics[width=\columnwidth]{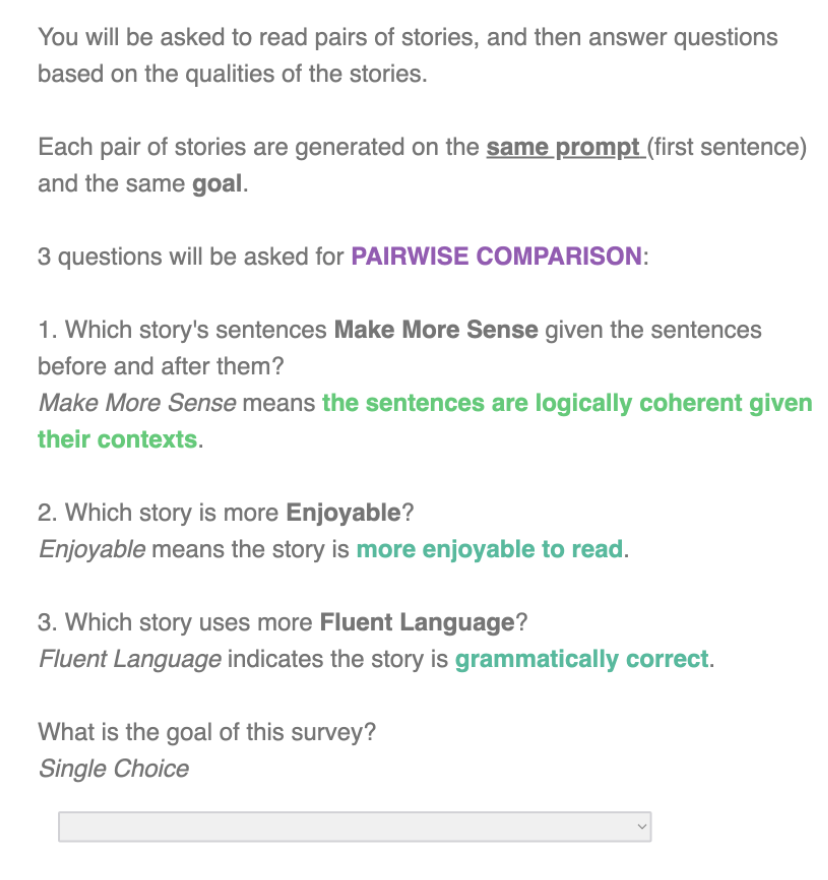}
    \caption{Screenshot of the human study instruction.}
    \label{fig:coherence_ins}
\end{figure}

\begin{figure}[!tbh]
    \centering
    \includegraphics[width=\columnwidth]{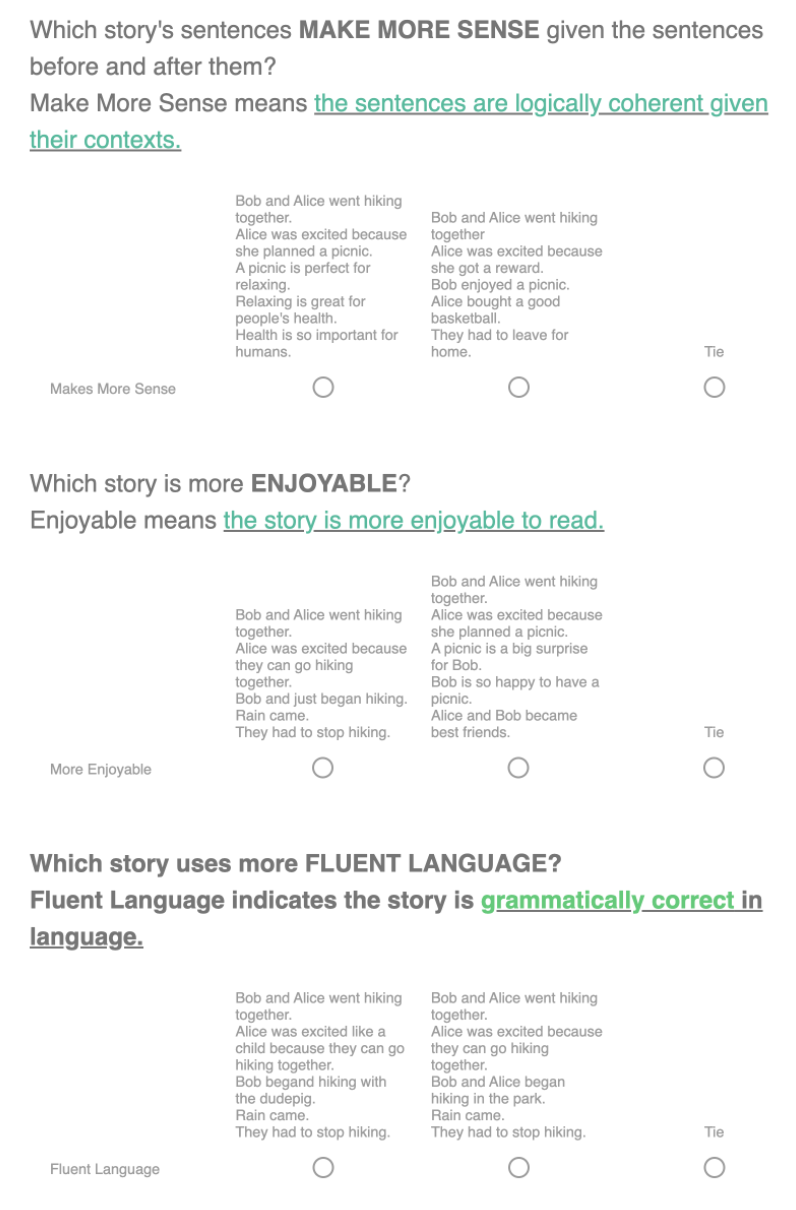}
    \caption{Screenshot of the human study screen questions.}
    \label{fig:coherence_s}
\end{figure}

\begin{figure}[!tbh]
    \centering
    \includegraphics[width=\columnwidth]{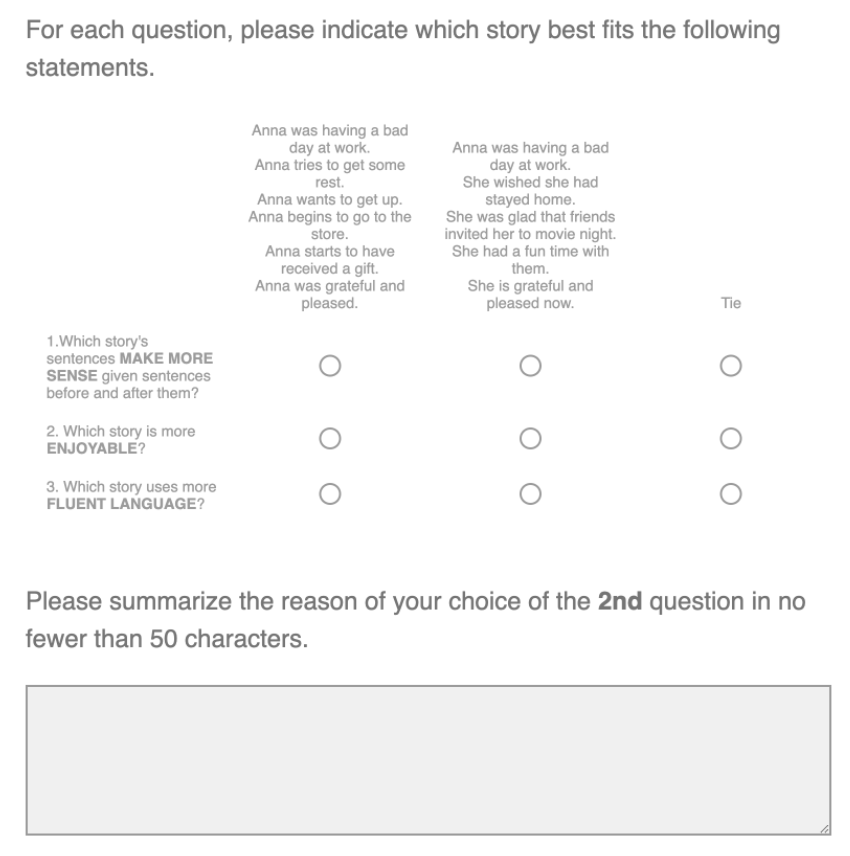}
    \caption{Screenshot of the human study on evaluating coherence.}
    \label{fig:coherence_exp_f}
\end{figure}

We evaluate coherence using human participant evaluation, asking a set of questions that includes dimensions such a logical coherence, loyalty to plot, and enjoyability. 
We recruited $50$ participants on a crowdsourcing platform.
We first show them the instruction( Figure~\ref{fig:coherence_ins}) and the screen questions (Figure~\ref{fig:coherence_s}).
We then ask questions in Section~\ref{sec:coherence_exp} and example is shown in Figure~\ref{fig:coherence_exp_f}.

We conduct our studies using the Cloud Research crowdsourcing platform to interact with Amazon Mechanical Turk \cite{litman2017turkprime}. 
Obtaining at least a bachelor's degree and English as their native language are required to take this study.
Participants are required to pass screening questions and then explain their preferences  of each choice in this study with more than $50$ characters,
which helps filter out low-quality responses and ensures the validity of the study. 
Our study was approved by our Institutional Review Board, and we payed participants the equivalent of $\$15$/hr.

\subsection{Controllability Evaluation Set-up}
\label{app:controllability_setup}
\begin{figure}[!tbh]
    \centering
    \includegraphics[width=\columnwidth]{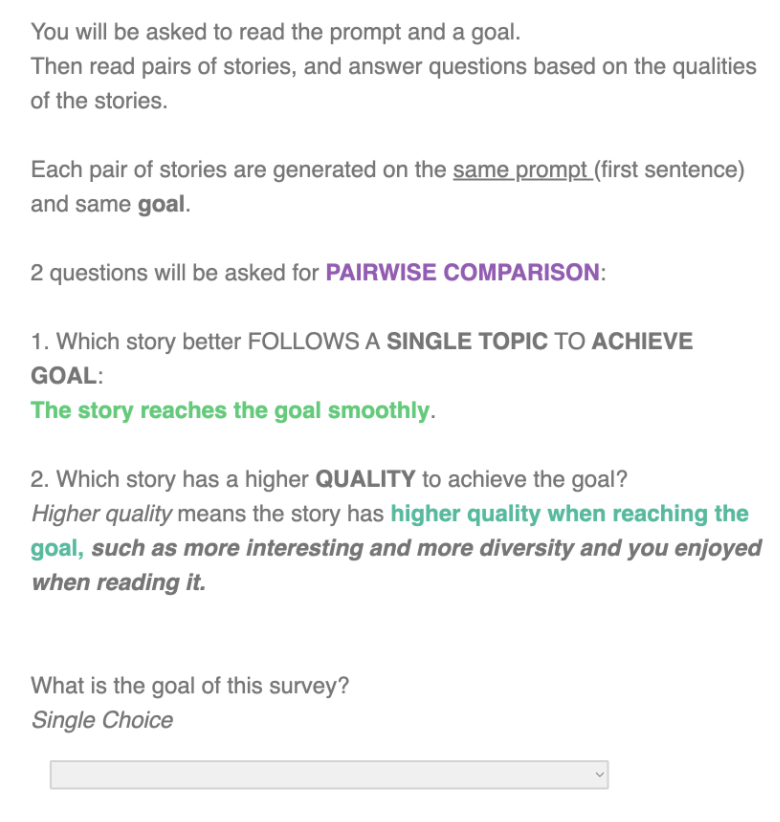}
    \caption{Screenshot of the instruction on evaluating coherence.}
    \label{fig:control_ins}
\end{figure}
 
 \begin{figure}[!tbh]
    \centering
    \includegraphics[width=\columnwidth]{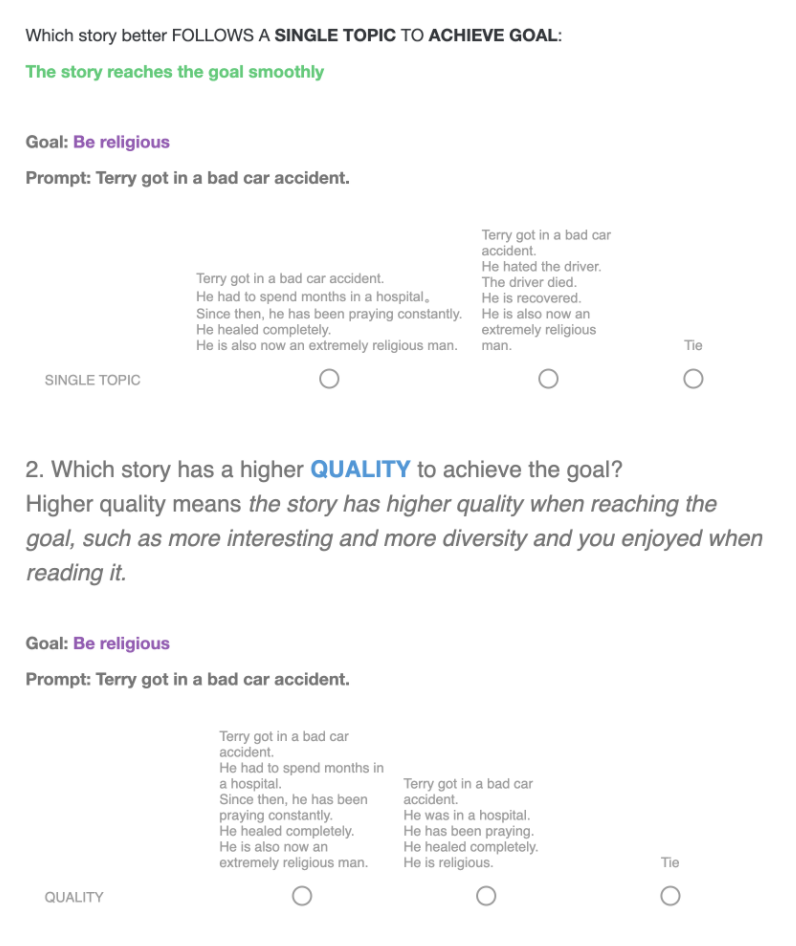}
    \caption{Screenshot of the screen questions on evaluating coherence.}
    \label{fig:control_scr}
\end{figure}
 
 \begin{figure}[!tbh]
    \centering
    \includegraphics[width=\columnwidth]{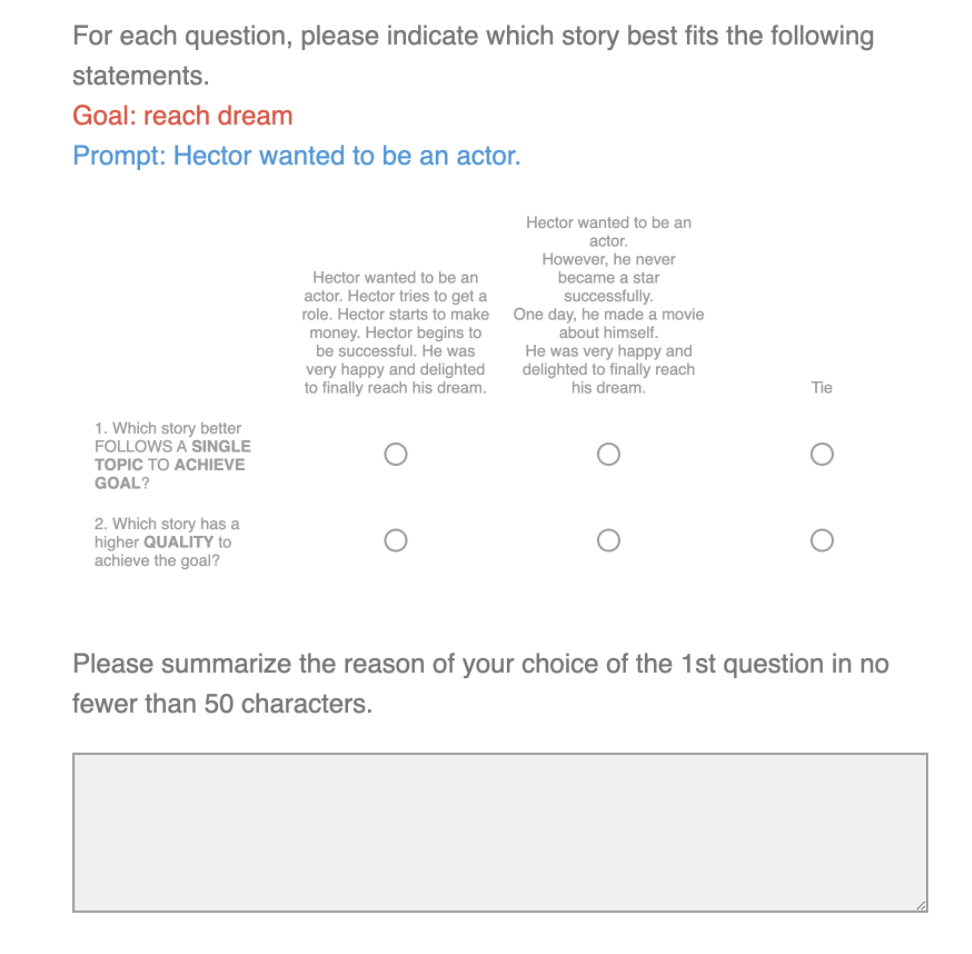}
    \caption{Screenshot of the human study on evaluating coherence.}
    \label{fig:control_exp_app}
\end{figure}

We recruited $48$ participants on a crowdsourcing platform.
We firstly show human participants instructions (Figure~\ref{fig:control_ins}) and screen questions (Figure~\ref{fig:control_scr}) to make sure they understand the task. 
We then ask them to  read a randomly selected generated story pairs---one from \sysname{} and one from our baselines.
They then answered which one better met the criteria in Section~\ref{sec:control_exp}.
Example is shown in Figure~\ref{fig:control_exp_app}.

We conduct our studies using the Cloud Research crowdsourcing platform to interact with Amazon Mechanical Turk \cite{litman2017turkprime}. 
Obtaining at least a bachelor's degree and English as their native language are required to take this study.
Participants are required to pass screening questions and then explain their preferences  of each choice in this study with more than $50$ characters,
which helps filter out low-quality responses and ensures the validity of the study. 
Our study was approved by our Institutional Review Board, and we payed participants the equivalent of $\$15$/hr.

\clearpage
\section{Evaluation Results}
\label{app:eva_results}
\subsection{Story Coherence Evaluation}
\label{app:coherence_app_results}
We also report the majority vote and agreement results in Table~\ref{tab:coherence_majority}.
We additionally observe that these three metrics are positively correlated using Spearman’s Rank Order Correlation in all of these ablation studies. $r_s = 0.49$, $p < 0.01$, between ``Logical Sense'' and ``Enjoyable''; 
$r_s = 0.39$, $p < 0.01$, between ``Logical Sense'' and ``Fluency'' for comparing to C2PO on ROCStories.
$r_s = 0.32$, $p < 0.01$, between ``Logical Sense'' and ``Enjoyable''; 
$r_s = 0.42$, $p < 0.01$, between ``Logical Sense'' and ``Fluency'' and  
$r_s = 0.32$, $p < 0.01$, between ``Enjoyable'' and ``Fluency'' for comparing to \citeauthor{goldfarb2020content} on Writing Prompts.
$r_s = 0.52$, $p < 0.01$, between ``Logical Sense'' and ``Enjoyable''; 
$r_s = 0.49$, $p < 0.01$, between ``Logical Sense'' and ``Fluency'' and  
$r_s = 0.37$, $p < 0.01$, between ``Enjoyable'' and ``Fluency'' 
for comparing to C2PO on fairy tale story dataset.

\begin{table*}[!tbh]
\footnotesize
\centering
\setlength\tabcolsep{5pt} %
\begin{tabular}{c|c|lll|lll|lll}
    \toprule
    \multirow{2}{*}{\textbf{Models}} & \textbf{Data}&
     \multicolumn{3}{c|}{\textbf{Logical Sense }} & \multicolumn{3}{c|}{\textbf{Enjoyable}} & \multicolumn{3}{c}{\textbf{Fluency}}\\
    
    & \textbf{set} 
    &Win\%&Lose\%&Tie\%&Win\%&Lose\%&Tie\%&Win\%&Lose\%& Tie\%
    \\

    \midrule
    \sysname{} vs CP
    &WP
      & \textbf{86.6}**$\dagger$ & 6.7 & 6.7 &  
      \textbf{60.0} & 40.0  & 0.0   
      & \textbf{80.0}** & \textbf{10.0} & 10.0
      
       \\
    \midrule
    
    \multirow{2}{*}{\sysname{} vs C2PO}
    &ROC
      & \textbf{80.0}**$\dagger$ & 10.0 & 10,0 &  
      \textbf{80.0}** & 10.0  & 10.0  
      & \textbf{60,0}**$\dagger$ & \textbf{0,0} & 40.0
       \\
    
    &FT
      & \textbf{60.0}* & 33.3 & 6.7 &  \textbf{73.3}** & 16.7  & 10.0
      & \textbf{60.0}$\dagger$& 20.0 & 20.0
       \\
    \bottomrule
\end{tabular}
\caption{Coherence evaluation results, 
showing the majority vote of participants who preferred the first system, second system, or thought the systems were equal. CP indicates \citep{goldfarb2020content}.
Each system is conditioned on the same test-set prompts and same goal. * indicates results are significant at $p<0.05$ confidence level; ** at $p<0.01$ using a Wilcoxan sign test on win-lose pairs. 
$\dagger$ indicates $\kappa$ > 0.2 or fair agreement. 
}
\label{tab:coherence_majority}
\end{table*}

\subsection{Story Controllability Evaluation}
\label{app:control_app_results}
We also report the majority vote and agreement results in Table~\ref{tab:Controllability_majority}.
We additionally observe that these two metrics are positively correlated using Spearman’s Rank Order Correlation in all of these ablation studies.
$r_s = 0.42$, $p < 0.01$, between ``goal'' and ``quality'' for comparing to C2PO on ROCStories;
$r_s = 0.39$, $p < 0.01$, between ``goal'' and ``quality'' for comparing to \citeauthor{goldfarb2020content} on Writing Prompts;
and $r_s = 0.39$, $p < 0.01$, between ``goal'' and ``quality'' for comparing to C2PO on fairy tale stories.

\begin{table*}[!thb]
\footnotesize
\centering
\setlength\tabcolsep{5pt} %
\begin{tabular}{c|c|lll|lll}
    \toprule
    \multirow{2}{*}{\textbf{Models}} & \textbf{Data}&
     \multicolumn{3}{c|}{\textbf{Goal}} & \multicolumn{3}{c}{\textbf{Quality}}\\
    
    &\textbf{set} & Win\%&Lose\%&Tie\%&Win\%&Lose\%&Tie\%\\

    \midrule

    \sysname{} vs CP
    &WP
      & \textbf{60.0}*$\dagger$ & 6.7 & 33.3 &  \textbf{53.3}** & 0.0  & 46.7
      \\
    \midrule
    
    \multirow{2}{*}{\sysname{} vs C2PO}
    &ROC
      & \textbf{86.6}**$\dagger$ & 6.7  & 6.7 &  \textbf{86.7}** & 0.0 & 13.3     
      \\
    
    &FT&
    \textbf{60.0}* & 6.7  & 33.3  
      & \textbf{60.0}* & 6.7 & 33.3 
      \\
    \bottomrule
\end{tabular}
\caption{Controllability evaluation results, 
showing the majority vote of participants who preferred the first system, second system, or thought the systems were equal.
CP indicates \citeauthor{goldfarb2020content}\shortcite{goldfarb2020content}.
Each system is conditioned on the same test-set prompts. * indicates results are significant at $p<0.05$ confidence level; ** at $p<0.01$ using a Wilcoxan sign test on win-lose pairs. $\dagger$ indicates $\kappa$ > 0.2 or fair agreement. 
}
\label{tab:Controllability_majority}
\end{table*}

\clearpage
\section{Ablation Study}
We perform ablation studies to choose the best hyperparameters,
We build goal story world states (\S\ref{sec:kg}) on the last sentence of randomly selected $30$ stories from ROCStories \cite{mostafazadeh2016corpus} to guide the story generation process.
\sysname{} keeps generating story continuations until knowledge graph difference score $R(s)$ reaches $0.8$ or generated story hits the goal ($r_1(s) = 1$). 
We measure the following metric:
\begin{itemize}[noitemsep,topsep=3pt,itemsep=3pt,leftmargin=*]
    \item \textit{Average story length} (Avg. len): Calculate the average story length which is required to reach $R(s) = 0.8$ (\S\ref{sec:graph}) or generated story hits the goal ($r_1(s) = 1$).
    Smaller average story length stands for faster, and thus more direct, goal achievement. 
    Because the system finds different ways to achieve the goal. As with many planning systems, ours has a bias toward shorter, more compact solutions.
    We stop generation when story length reaches $10$.
    
\end{itemize}

\begin{table}[h]
\footnotesize
\centering
\begin{tabular}{c|c|l}
\toprule
\multicolumn{2}{c}{\textbf{Model}} & \textbf{Avg. len} $\downarrow$  \\
\midrule
\multirow{3}{*}{\shortstack{\sysname\\ Full}}& $\alpha=0.50$ &  ${\textbf{6.92} \pm 1.21}$  \\
& $\alpha=0.90$ & ${9.27 \pm 1.50}$  \\
& $\alpha=0.25$ &  ${8.32 \pm 0.94}$  \\ 
\bottomrule
\end{tabular}
\caption{Results of the ablation study. 
$\alpha$ is tuning the inference contribution when calculating graph difference. 
}
\label{tab:ablation}
\end{table}

Table~\ref{tab:ablation} shows the result of the ablation study. 
We experiment with three values of $\alpha$ in our \sysname{} framework.
The best performing model has $\alpha=0.5$, balancing between inference-node-guided and goal-node-guided story generation, where larger $\alpha$ indicates more inference-node-driven.

\end{document}